\pdfoutput=1

\documentclass[11pt]{article}

\usepackage[]{ACL2023}

\usepackage{times}
\usepackage{latexsym}

\usepackage[T1]{fontenc}

\usepackage[utf8]{inputenc}

\usepackage{microtype}

\usepackage{inconsolata}

\title{Ambiguity Meets Uncertainty: Investigating Uncertainty Estimation for Word Sense Disambiguation}

\author{Zhu Liu\\
  Tsinghua University  \\
  School of Humanities\\
  \texttt{liuzhu22@mails.tsinghua.edu.cn} \\\And
  Ying Liu \\
  Tsinghua University  \\
  School of Humanities\\
  \texttt{yingliu@tsinghua.edu.cn} \\}

\usepackage{booktabs}
\usepackage{amsfonts}
\usepackage{amsmath}
\usepackage{color,xcolor} 
\usepackage{xspace}
\usepackage{adjustbox}
\usepackage{multirow}
\newcommand*{\eg}{e.g.\@\xspace}
\newcommand*{\ie}{i.e.\@\xspace}

\usepackage{amssymb}
\usepackage{algpseudocode}
\usepackage{xcolor,colortbl}
\usepackage{bbm}
\usepackage{hyperref}

\makeatletter

\newcommand{\Rmnum}[1]{\mathrm{\expandafter\@slowromancap\romannumeral #1@}}
\makeatother

\begin{document}

\setcounter{page}{1}

\maketitle
\begin{abstract}
Word sense disambiguation (WSD), which aims to determine an appropriate sense for a target word given its context, is crucial for natural language understanding. Existing supervised methods treat WSD as a classification task and have achieved remarkable performance. However, they ignore uncertainty estimation (UE) in the real-world setting, where the data is always noisy and out of distribution. This paper extensively studies UE on the benchmark designed for WSD. Specifically, we first compare four uncertainty scores for a state-of-the-art WSD model and verify that the conventional predictive probabilities obtained at the final layer of the model are inadequate to quantify uncertainty. Then, we examine the capability of capturing data and model uncertainties by the model with the selected UE score on well-designed test scenarios and discover that the model adequately reflects data uncertainty but underestimates model uncertainty. 
Furthermore, we explore numerous lexical properties that intrinsically affect data uncertainty and provide a detailed analysis of four critical aspects: the syntactic category, morphology, sense granularity, and semantic relations. The code is available at \url{https://github.com/RyanLiut/WSD-UE}.

\end{abstract}

\section{Introduction}
Disambiguating a word in a given context is fundamental to natural language understanding (NLU) tasks, such as machine translation \cite{MT}, question answering \cite{QA}, and coreference resolution \cite{CR}. This task of word sense disambiguation (WSD) targets polysemous or homonymous words and determines the most appropriate sense based on their surrounding contexts. For example, the ambiguous word \textit{book} refers to two completely distinct meanings in the following sentences: i)``\textit{Book} a hotel, please.'', ii) ``Read the \textit{book}, please''. The phenomenon is universal to all languages and has been paid much attention since the very beginning of artificial intelligence (AI) \cite{weaver1952translation}.

\begin{figure}
    \centering
    \includegraphics[width=1.0\linewidth]{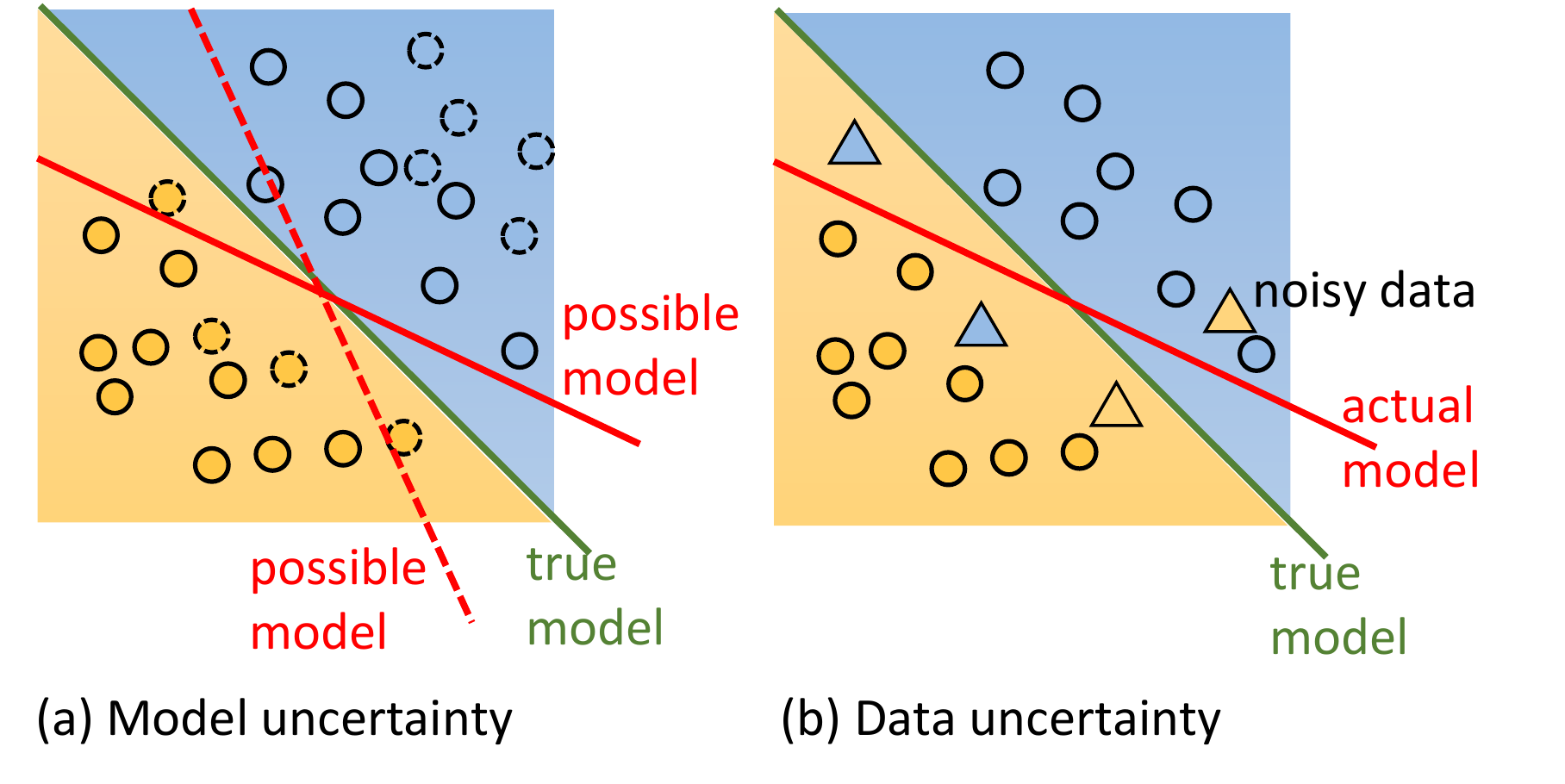}
    \caption{Two types of uncertainties in the case of classification. The green line indicates the true model (decision boundary), while the red shows possible models. Circles and triangles with different colors illustrate clean and noisy data with corresponding labels.}
    \label{fig:UEDE}
\end{figure}

Existing supervised methods \cite{BEM, ML-WSD, EWISER, EViLBERT, GBERT} cast WSD as a classification task in which a neural networks (NNs)-based classifier is trained from WordNet \cite{WordNet}, a dictionary-like inventory. Although they have achieved the state of the art on WSD benchmarks, with some even breaking through the estimated upper bound on human inter-annotator agreement in terms of accuracy \cite{EWISER}, they do not capture or measure uncertainty. Uncertainty estimation (UE) answers a question as follows: \textit{To what extent is the model certain that its choices are correct?} A model can be unsure due to the noisy or out-of-domain data, especially in a real-world setting. This estimation delivers valuable insights to the WSD practitioners since we could pass the input with high uncertainty to a human for classification.

UE is an essential requirement for WSD. Interestingly, the word ``ambiguous'' (in terms of the task of word sense \textit{disambiguation}) itself is ambiguous: it refers to i) doubtful or uncertain especially from obscurity or indistinctness, and ii) capable of being understood in two or more possible senses or ways, according to the Merriam-Webster dictionary\footnote{\url{https://www.merriam-webster.com/dictionary/ambiguous}}. The conventional treatment only considers its second aspect but disregards the first uncertainty-related sense. In reality, there are many situations where uncertainties arise \cite{galthesis}. The first situation assumes a true model to which each trained model approximates. Uncertainty appears when the structures and parameters of the possible models vary; we refer to it as model uncertainty (Figure~\ref{fig:UEDE} (a)) in this paper. \textit{Model uncertainty} can be reduced when collecting enough data, \ie, adequate knowledge to recognize the true model and out-of-distribution (OOD) data is always used to test model uncertainty. It has been observed that WSD is prone to domain shift and bias towards the most frequent sense (MFS) \cite{evaluation}. Therefore, it is essential to quantify model uncertainty in the task.

Another uncertainty is related to the data itself and cannot be explained away, which is referred to as \textit{data uncertainty} (also called aleatoric uncertainty). Data uncertainty happens when the observation is imperfect, noisy, or obscure (Figure~\ref{fig:UEDE} (b)). Even if there is enough data, we cannot obtain results with high confidence. WSD is context-sensitive, and the model output could be divergent due to partial or missing context. Even worse, some words have literal and non-literal meanings and can be understood differently. With a fine-grained WordNet \cite{WordNet} as a reference inventory, the inter-annotator disagreement is up to 20\% to 30\% \cite{2009survey}: even human annotators cannot agree on the correct sense of these words.

In this paper, we perform extensive experiments to assess the uncertainty of a SOTA model \cite{ML-WSD} on WSD benchmarks. First, we compare the probability of the model output with the other three uncertainty scores and conclude that this probability is inadequate to UE, which is consistent with previous research \cite{gal2016dropout}. Then, with the selected score, we evaluate data uncertainty in two designed scenarios: window-controlled and syntax-controlled contexts, which simulate noisy real-world data. Further, we estimate model uncertainty on an existing OOD dataset \cite{42D} and find that the model underestimates model uncertainty compared to the adequate measure of data uncertainty. Finally, we design an extensive controlled procedure to determine which lexical properties affect uncertainty estimation. The results demonstrate that morphology (parts of speech and number of morphemes), inventory organization (number of annotated ground-truth senses and polysemy degree) and semantic relations (hyponym) influence the uncertainty scores.

\section{Related Work}
\subsection{Word Sense Disambiguation}
Methods of WSD are usually split into two categories, which are knowledge-based and supervised models. Knowledge-based methods employ graph algorithms, e.g., clique approximation \cite{Babelfy}, random walks \cite{UKB}, or game theory \cite{GT} on semantic networks, such as WordNet \cite{WordNet}, BabelNet \cite{BabelNet}. These methods do not acquire much annotation effort but usually perform worse than their supervised counterpart due to their independence from the annotated data. Supervised disambiguation is data-driven and utilizes manually sense-annotated data sets. 
Regarding each candidate sense as a class, these models treat WSD as the task of multi-class classification and utilize deep learning techniques, \eg, transformers \cite{ML-WSD,bevilacqua2019quasi}. Some also integrate various parts of the knowledge base, such as neighboring embeddings \cite{loureiro2019language}, relations \cite{ML-WSD}, and graph structure \cite{EWISER}. These methods have achieved SOTA performance and even broken through the ceiling human could reach \cite{EWISER}. 
However, these methods treat disambiguation as a deterministic process and neglect the aspect of uncertainty.

\subsection{Uncertainty Estimation}
Uncertainty estimation (UE) has been studied extensively, especially in computer vision \cite{gal2017deep} and robust AI \cite{stutz2022understanding}. Methods capture uncertainty in a Bayesian or non-Bayesian manner. Bayesian neural networks \cite{neal2012bayesian} offer a mathematical grounded framework to model predictive uncertainty but usually comes with prohibitive inference cost. Recent work proved MC Dropout approximates Bayesian inference in deep Gaussian Processes and has been widely applied in many UE applications \cite{vazhentsev2022uncertainty, kochkina2020estimating} due to its simplicity. Other non-Bayesian approaches include Maximum Softmax Probability \cite{MP}, Label Smoothing \cite{LabelSmooth}, Calibration \cite{guo2017calibration}, etc., in the context of selective prediction \cite{varshney2022towards}. During recent years, the field of natural language processing has witnessed the development of an increasing number of uncertain-aware applications, such as Machine Translation \cite{glushkova2021uncertainty}, Summarization \cite{gidiotis2021uncertainty}, Question Answering \cite{varshney2023post} and Information Retrieval \cite{penha2021calibration}. Nevertheless, little attention has been paid to the combination of UE and WSD. An early work \cite{zhu2008active} explored uncertainty to select informative data in their active learning framework. However, the uncertainty estimation for WSD is not explored extensively, as we do in a quantitative and qualitative way.

\section{Uncertainty Scenarios}
\subsection{Problem Formulation}
Given a target word $w_i$ in a context $c_i=(w_{0},w_{1},...,w_{i},...,w_{W})$ of $W$ words, a WSD model selects the best label $\hat{y}_i$ from a candidate sense set $S_i=(y_1, y_2, ..., y_M)$ consisting of $M$ classes. A neural network $p_\theta$ with the parameter $\theta$ usually obtains a probability $p_i$ over $M$ classes by a softmax function which normalizes the model output $f_i$:
\begin{equation}
    p_i = \text{SoftMax} (f_i(w_i|c_i;\theta)).
\end{equation}
During training, the probability is used to calculate cross-entropy loss, which can be recognized as a probability for each candidate class during the inference. Such a point estimation of model function has been erroneously interpreted as model confidence \cite{gal2016dropout}. The goal of UE is to find a suitable $p_i$ to better reflect true predictive distribution under data and model uncertainty sources. Suppose we have a reasonable score $s(p_i) \in \mathcal{S}$ indicating UE, where $\mathcal{S}$ is a metric space, we expect $s^a > s^b$ when a situation $a$ is more uncertain than $b$.

\subsection{Data Uncertainty: Controllable Context}
\begin{figure}[h!]
    \centering
    \includegraphics[width=1.0\linewidth]{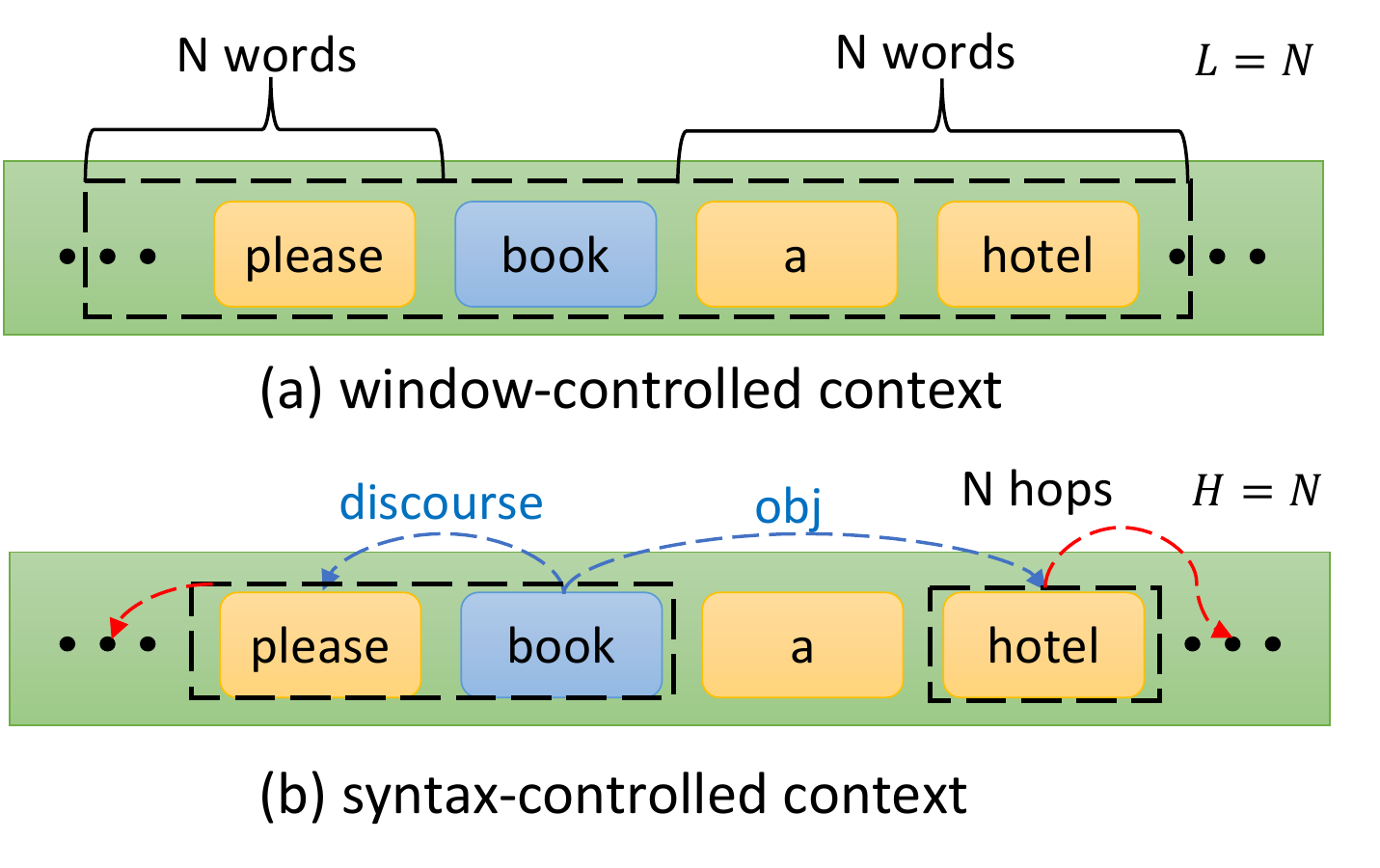}
    \caption{Two types of controlled context in the data uncertainty setting. The target word is highlighted in blue. The box with a black dotted line shows the final chosen context. We show the dependency relation in \textcolor{blue}{blue} and \textcolor{red}{red}.}
    \label{fig:context}
\end{figure}

Data uncertainty measures the uncertainty caused by imperfect or noisy data. We consider that such noises could happen in the context surrounding the target word, considering WSD is a context-sensitive task. With different degrees of missing parts in the context, the model is expected to obtain predictions with different qualifications of uncertainty. To simulate this scenario, we control the range of context based on two signals: the window and the syntax, as illustrated in Figure~\ref{fig:context}.

\subsubsection{Window-controlled Context}
We choose $L$ words both on the left and right of the target word $w_i$ as the window-controlled context $c_L^{\text{WC}}=(w_{l},w_{i-1},w_{i}, w_{i+1},...,w_{h})$, where $l=\max (i-L, 0)$ and $h=\min (i+L, W)$ are the lower index and the higher index. With a hypothesis that longer context tends to contain more clues to disambiguate a word and a suitable UE score $s$, we expect that $s^{\text{WC}}_a > s^{\text{WC}}_b$, where two window-controlled contexts are extracted with the length of $a$ and $b$, and $a < b$. 

\subsubsection{Syntax-controlled Context}
In our second controlled method, we utilize the neighboring syntax around $w_i$. Specifically, we parse the universal syntactical dependency relations between words using tools of Stanza \cite{stanza}. This is represented as a form of graph structure $\mathcal{G}=(\mathcal{N}, \mathcal{R})$, where $\mathcal{N}$ denotes the nodes, \ie, each word, and $\mathcal{R}=<n^h, n^t, r>$ is the relation $r$ from the head node $n^h$ to tail node $n^t$. For example, when $r$ is \textit{nsubj}, that means $n^h$ is the subject of $n^t$. We iteratively obtain a syntax-related \footnote{We denote this scenario as DP, since we utilize dependency parsing as the syntactic representation.} neighboring set with the $H$ hops of the target word $w_i$ as $c_H^{\text{DP}}$ in the following approach. Initially, $c_H^{\text{DP}}$ only contains $w_i$. After one hop, $c_H^{\text{DP}}$ collects the head node and tail nodes of $w_i$. The procedure is repeated $H$ times, with more syntactically related words added. We also rationally hypothesize a smaller $s^{\text{DP}}$, which measures uncertainty under syntax-controlled context, favors the context with a larger $H$. We highlight that the syntax-controlled context leverages the nonlinear dependency distance \cite{heringer1980syntax} between words in connection, compared to the linear distance in the scenario of window-controlled context. 

\subsection{Model Uncertainty: OOD Test}
Model uncertainty is another crucial aspect of UE, widely studied in the machine learning community. Lacking knowledge, models with different architectures and parameters could output indeterminate results. Testing a model on OOD datasets is a usual method to estimate model uncertainty. In the task of WSD, we employ an existing dataset 42D \cite{42D} designed for a more challenging benchmark. This dataset built on the British National Corpus is challenging because 1) for each instance, the ground truth does not occur in SemCor \cite{semcor}, which is the standard training data for WSD, and 2) is not the first sense in WordNet to avoid most frequent sense bias issue \cite{DIBIMT}. 42D also has different text domains from the training corpus. These confirm that 42D is an ideal OOD dataset.

\section{Experiments}

\begin{table*}[t]
\small
    \begin{center}
    \begin{tabular}{c|cc|cc|cc|cc|cc}
    \toprule
    \multirow{2}{*}{UE Score} & \multicolumn{2}{c|}{Senseval-2} & \multicolumn{2}{c|}{Senseval-3} & \multicolumn{2}{c|}{SemEval-07} & \multicolumn{2}{c|}{SemEval-13} & \multicolumn{2}{c}{SemEval-15} \\
     & RCC $\downarrow$ & RPP $\downarrow$ & RCC $\downarrow$ & RPP $\downarrow$ & RCC $\downarrow$ & RPP $\downarrow$  & RCC $\downarrow$ & RPP $\downarrow$ & RCC $\downarrow$ & RPP $\downarrow$\\ 
    
    \midrule
        MP&\textbf{5.69}&9.50&7.11&10.37&\textbf{8.68}&11.40&5.78&8.02&\textbf{5.02}&\textbf{11.07} \\
 \rowcolor[gray]{0.8} SMP&5.78&\textbf{9.14}&\textbf{7.10}&\textbf{9.83}&8.81&\textbf{10.83}&\textbf{5.59}&\textbf{7.88}&5.34&11.16 \\
PV&6.11&11.47&7.50&12.40&9.93&16.00&5.97&10.22&5.62&13.11 \\ 
BALD&6.00&11.09&7.46&11.99&9.36&14.73&5.83&10.02&5.48&12.77 \\
        \bottomrule
          
    \end{tabular}
\end{center}
    \caption{UE score comparisons on five standard WSD datasets.}
    \label{tab:four}
\end{table*}

\begin{table*}[t]
\small
    \begin{center}
    \begin{tabular}{c|cc|cc|cc|cc|cc}
    \toprule
    \multirow{2}{*}{UE Score} & \multicolumn{2}{c|}{NOUN} & \multicolumn{2}{c|}{VERB} & \multicolumn{2}{c|}{ADJ} & \multicolumn{2}{c|}{ADV} & \multicolumn{2}{c}{ALL} \\
     & RCC $\downarrow$ & RPP $\downarrow$ & RCC $\downarrow$ & RPP $\downarrow$ & RCC $\downarrow$ & RPP $\downarrow$  & RCC $\downarrow$ & RPP $\downarrow$ & RCC $\downarrow$ & RPP $\downarrow$ \\ 
    
    \midrule
MP&6.06&\textbf{7.47}&14.08&18.20&5.15&\textbf{8.25}&3.70&4.89&6.13&9.78 \\
 \rowcolor[gray]{0.8} SMP&\textbf{4.94}&7.66&\textbf{13.76}&\textbf{17.45}&\textbf{4.39}&8.35&\textbf{2.65}&\textbf{4.85}&\textbf{6.11}&\textbf{9.44} \\
PV&6.25&9.17&15.38&22.02&4.97&9.37&3.20&5.33&6.48&11.91 \\
BALD&5.18&9.39&14.42&20.96&4.59&9.80&2.66&5.56&6.36&11.52 \\
        \bottomrule
          
    \end{tabular}
\end{center}
    \caption{UE score comparisons on all the datasets with different kinds of POS.}
    \label{tab:all}
\end{table*}

\subsection{Model and Datasets}
We conduct our UE for a SOTA model MLS \cite{ML-WSD}, with the best parameters released by the authors. They framed WSD as a multi-label problem and trained a BERT-large-cased model \cite{bert} on the standard WSD training dataset SemCor \cite{semcor}. We follow their settings except for using Dropout during inference when performing Monte Carlo Dropout (MC Dropout). We set the number of samples $T$ to be 20, conduct $3$ rounds, and report the averaged performance. 

As regards the evaluation benchmark, we use the Unified Evaluation Framework for English all-words WSD proposed by \cite{evaluation}. This includes five standard datasets, namely, Senseval-2, Senseval-3, SemEval-2007, SemEval-2013, and SemEval-2015. The whole datasets concatenating all these data with different parts of speech (POS) are also evaluated. Note that in our second part, We use a portion of SemEval-2007 to investigate data uncertainty and 42D is used for model uncertainty. 

\subsection{Uncertainty Estimation Scores}
We apply four methods as our uncertainty estimation (UE) scores. One trivial baseline \cite{geifman2017selective} regards the Softmax output $p_i$ as the confidence values over classes $y = s \in S$. We calculate the uncertainty score based on the maximum probability as $u_{\text{MP}}(x)=1-\max \limits_{s \in S} p(y=s|x)$. 

The other three methods are based on MC Dropout, which has been proved theoretically as approximate Bayesian inference in deep Gaussian processes \cite{gal2016dropout}. Specifically, we conduct $T$ stochastic forward passes during inference with Dropout random masks and obtain $T$ probabilities $p_t$. Following the work \cite{vazhentsev2022uncertainty}, we use the following measures:
\begin{itemize}
    \item Sampled maximum probability (SMP) takes the sample mean as the final confidence before an MP is applied: $u_{\mathrm{SMP}}=1-\max _{s \in S} \frac{1}{T} \sum_{t=1}^T p_t^s,$ where $p_t^s$ refers to the probability of belonging to class $s$ at the $t'th$ forward pass.
    \item Probability variance (PV) \cite{gal2017deep} calculates the variance before averaging over all the class probabilities: $u_{\mathrm{PV}}=\frac{1}{S} \sum_{s=1}^S\left(\frac{1}{T} \sum_{t=1}^T\left(p_t^s-\overline{p^s}\right)^2\right).$
    \item Bayesian active learning by disagreement (BALD) \cite{houlsby2011bayesian} measures the mutual information between model parameters and predictive distribution: $u_{\mathrm{BALD}}=-\sum_{s=1}^S \overline{p^s} \log \overline{p^s}+\frac{1}{T} \sum \limits_{s, t} p_t^s \log p_t^s.$
\end{itemize}

Note that these scores are instance-specific and we report the averaged results over all the samples.

\subsection{Metrics on UE scores}
While UE scores are a measure of uncertainty, we also need metrics to judge and compare the quality of different UE scores. A hypothesis is that a sample with a high uncertainty score is more likely to be erroneous and removing such instances could boost the performance. We employ two metrics following the work \cite{vazhentsev2022uncertainty}: area under the risk courage curve (\textbf{RCC}) \cite{RCC} and reversed pair proportion (\textbf{RPP}) \cite{RPP}. RCC calculates the cumulative sum of loss due to misclassification according to the uncertainty level for rejections of the predictions. A larger RCC indicates that uncertainty estimation negatively impacts the classification. Note that we use the normalized RCC by dividing the size of the dataset. RPP counts the proportion of instances whose uncertainty level is inconsistent with its loss level compared to another sample. For any pair of instances $x_i$ and $x_j$ with their UE score $u(x)$ and loss value $l(x)$:
\begin{equation}
\small
    RPP = \frac{1}{n^2} \sum _{i,j=1} ^{n} \mathbbm{1} [ u(x_i) < u(x_j), l(x_i)>l(x_j)],
\end{equation}
where $n$ is the size of the dataset.

\section{Results and Analysis}
In the first part, we show the quantitative results of different UE scores and the performances of data and model uncertainty. Then a qualitative result demonstrates specific instances with a range of uncertainties. This motivates us to analyze which lexical properties mainly affect uncertainty in the last part.

\subsection{Quantitative Results}
\label{Sec:QR}
\subsubsection{Which UE score is better?}
We measure the four UE scores, MP, SMP, PV, and BALD in terms of two metrics, RCC and RPP. The results of five standard datasets are shown in Table~\ref{tab:four} while the performance on all the datasets involving different parts of speech is demonstrated in Table~\ref{tab:all}. For most of the data, SMP outperforms the other three scores in spite of some inconsistent results where MP has a slight advantage, such as on SemEval-15. Interestingly enough, softmax-based scores \ie, MP and SMP, surpass the other two, PV and BALD. Similar results can be observed in the work \cite{vazhentsev2022uncertainty}. This may be due to the fact that the former scores are directly used as the input of the maximum likelihood objective, thus more accurately approximating the real distribution. 

\begin{figure}[h!]
    \centering
    \includegraphics[width=1\linewidth]{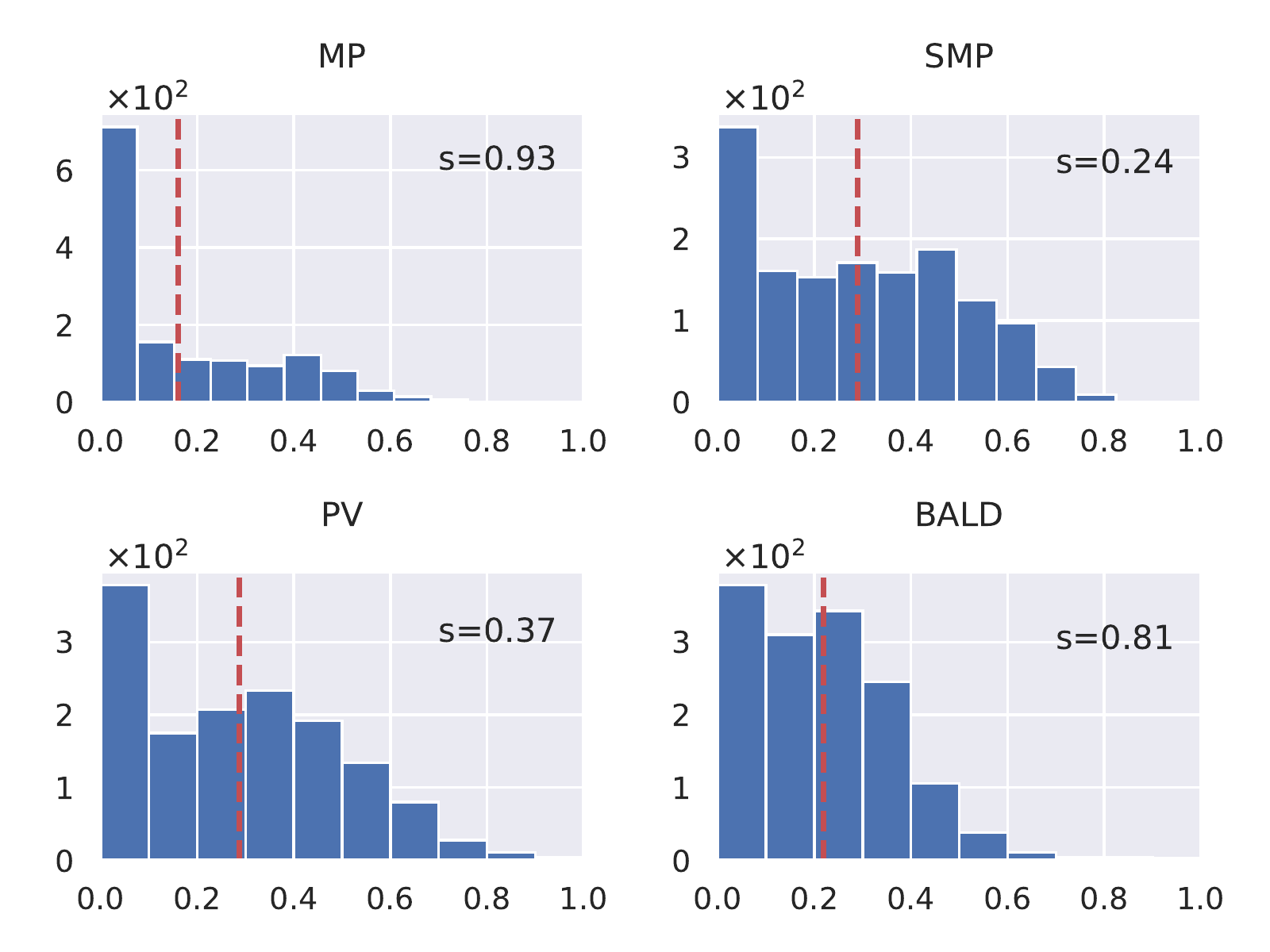}
    \caption{The distribution of four UE scores on misclassified instances of all datasets. A red dotted line indicates the average value. We calculate the sample skewness $s$ for each score as well. Note that PV and BALD scores are normalized into the range from 0 to 1.}
    \label{fig:dist}
\end{figure}

To further investigate the distribution of these four scores, we show the histograms of these scores in the misclassified instances, as illustrated in Figure~\ref{fig:dist}. We also display the averaged value (a red dotted line) and the sample skewness $s$, calculated as the Fisher-Pearson coefficient \citep{zwillinger1999crc}. Since here we focus on the misclassified samples, the cases of all the samples and those correctly classified are reported in Appendix~\ref{app:distribution}. This shows that MP has a more long-tailed and skewed distribution than scores based on MC Dropout, indicating MP is overconfident towards the wrong cases. However, the other three metrics have a more balanced distribution. This verifies the common concern on the SoftMax output of a single forward as an indication of confidence.

Finally, given its outstanding performance, we chose SMP as our uncertainty score in the following experiments. 

\subsubsection{How does the model capture data uncertainty?}

\begin{figure}[h!]
    \centering
    \includegraphics[width=1\linewidth]{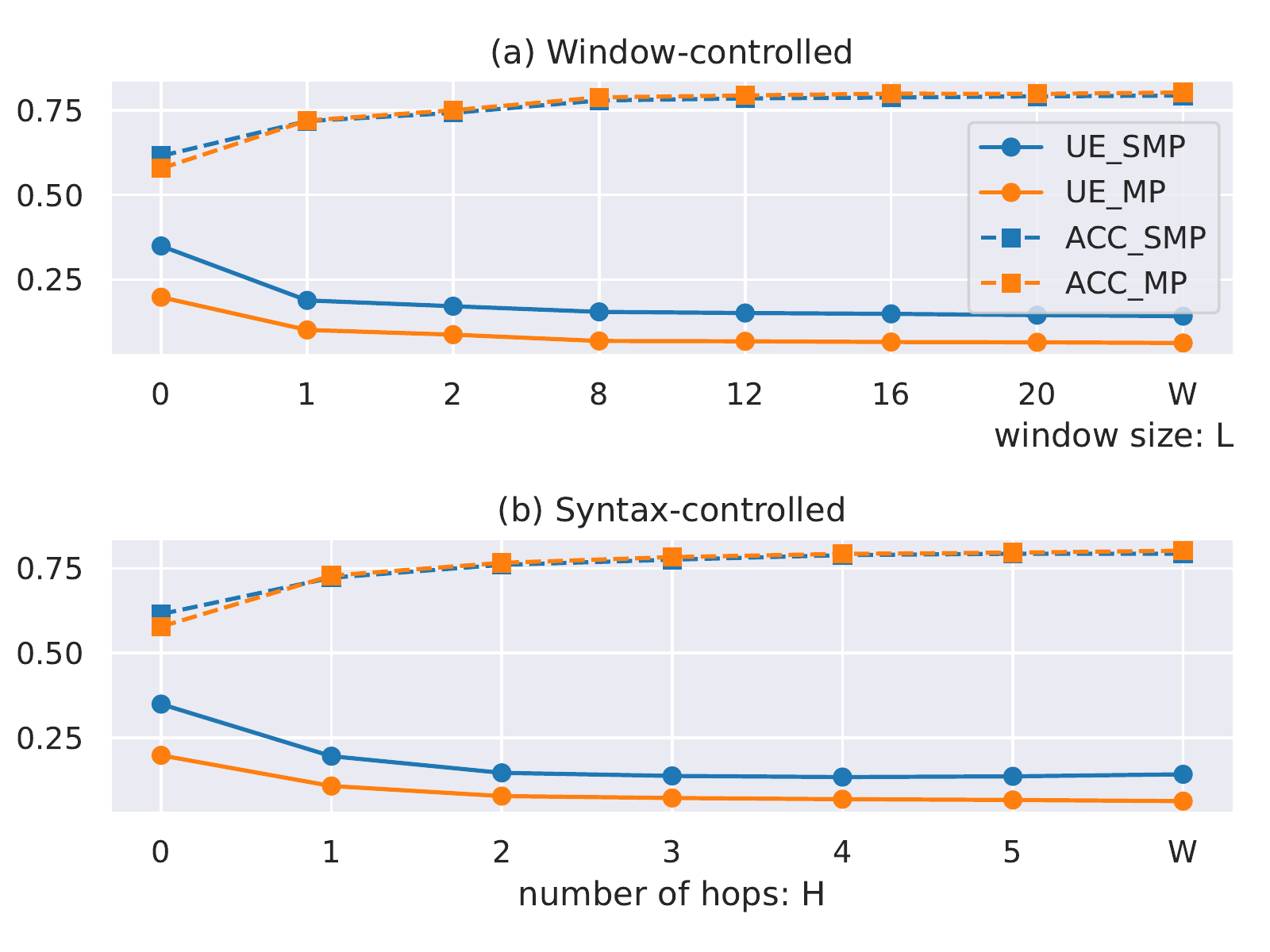}
    \caption{UE scores (SMP and MP) and accuracy (F1 score) vary depending on the range of context for (a) window-controlled setting and (b) syntax-controlled setting. Note that ``0'' indicates that only target words without context are available to the model. On the other hand, ``W'' means the whole context is available.}
    \label{fig:control}
\end{figure}

We verify data uncertainty in window-controlled and syntax-controlled scenarios, as shown in Figure~\ref{fig:control}. In the first setting, UE becomes less, and the accuracy grows with the increase of window size $T$. This indicates that the model perceives more and more confidence in the data, accessible to more neighboring words. The trend is similar in the syntax-controlled setting. These show that the model can adequately capture data uncertainty. SMP has a larger uncertainty than MP, especially in a sparse context, such as L or H is equal to 0 or 1, where the model is expected to be much more uncertain. We report the comparison of the other two sample-based scores, PV and BALD in Appendix~\ref{app:du}.

\subsubsection{How does the model capture model uncertainty?}

\begin{figure}[h!]
    \centering
    \includegraphics[width=1\linewidth]{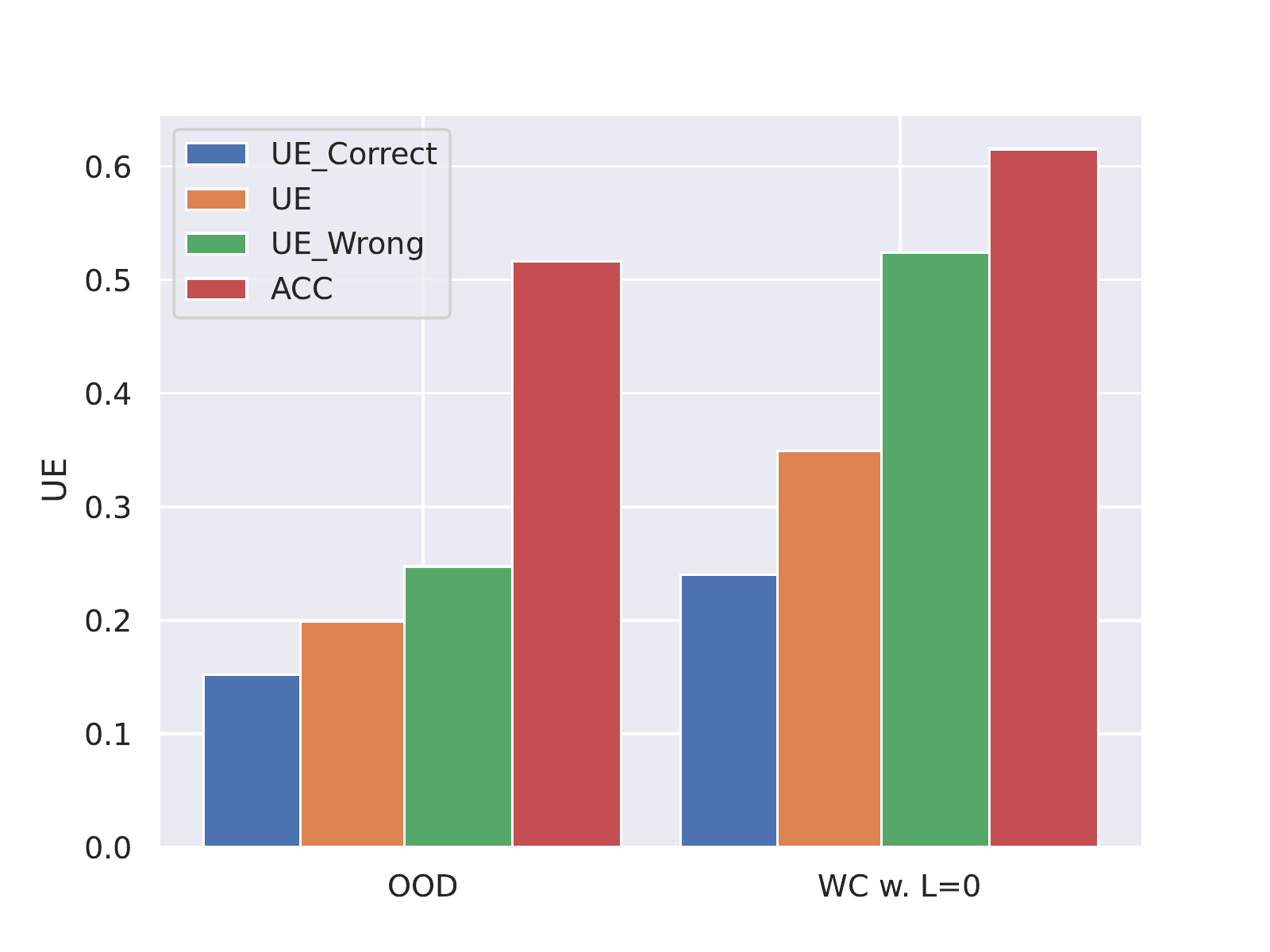}
    \caption{Uncertainty and accuracy (F1) scores for model uncertainty (OOD) and data uncertainty (controlled context) scenarios. We use window-controlled UE with \textit{L}=0 (WC w. \textit{L}=0). It is evaluated in all the data instances and wrongly (UE\_Wrong) or correctly (UE\_Correct) classified instances.}
    \label{fig:MU}
\end{figure}

We examine the model uncertainty on the 42D dataset in Figure~\ref{fig:MU}. The result shows OOD dataset is indeed a challenging benchmark for WSD. However, even with worse performance, the model fails to give a high UE score. We compare it with the most uncertain cases but similar accuracy in the settings of data uncertainty, \ie, without any context when $L=0$. The OOD setting has a lower level of uncertainty, especially in the misclassified samples, even if it has degraded performance. This implies that the model underestimates the uncertainty level in model uncertainty. We show the performance of MP, PV, and BALD in Appendix~\ref{app:mu}.

\subsection{Qualitative Results}
\begin{figure}[h!]
    \centering
    \includegraphics[width=1.0\linewidth]{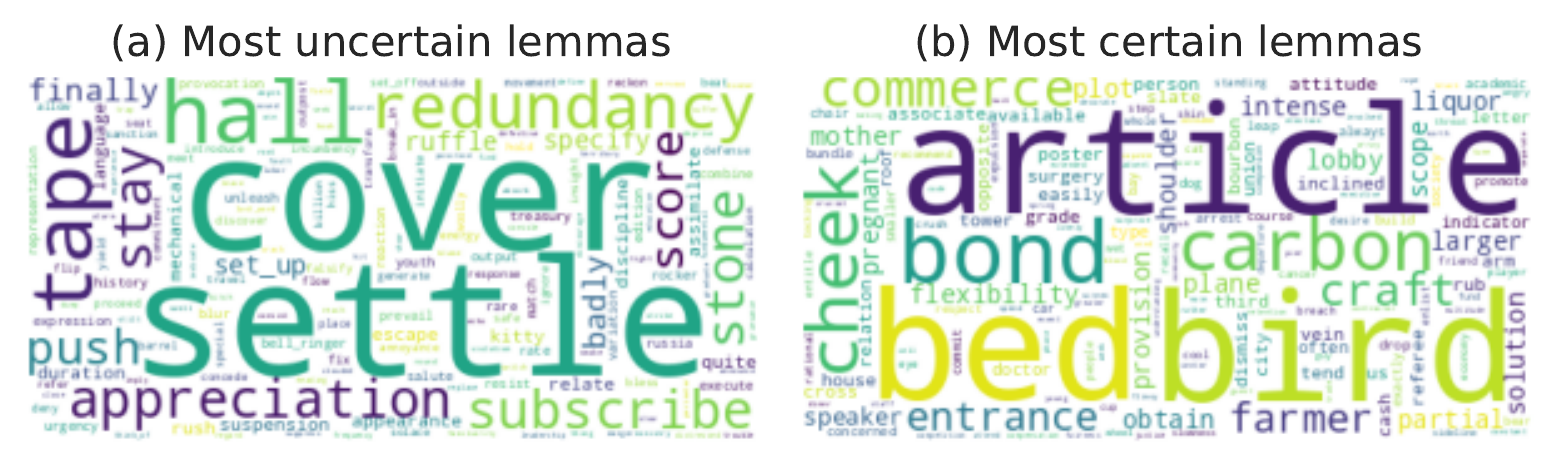}
    \caption{Word clouds for lemmas where a larger font indicates higher (a) or lower (b) UE scores. }
    \label{fig:effect_lemma}
\end{figure}

To investigate what kinds of words given a context tend to be uncertain, we obtain the final UE score for each word by averaging SMP scores for instances sharing the same form of lemma. In Figure~\ref{fig:effect_lemma}, We show the word clouds for words with the most uncertain (left (a)) and certain (right (b)) meanings. We remove some unrepresented words whose number of candidate senses is less than 3. With respect to the most uncertain lemmas, there are words such as \textit{settle}, \textit{cover} etc. Most of them are verbs and own multiple candidate senses. As for most certain cases, the senses of nouns like \textit{bird}, \textit{bed}, and \textit{article} are determined with low uncertainty. These phenomena motivate us to investigate which lexical properties affect uncertainty estimation in the next part. It is noted that we concentrate on data uncertainty instead of model uncertainty, based on the investigation in Subsection~\ref{Sec:QR}, which appears due to the data itself, \ie, lexical characteristics.

\subsection{Effects on Uncertainty} \label{sec:effects}
We explore which lexical properties affect uncertainty estimation from four aspects: the syntactic category \cite{folk2003effects}, morphology~\footnote{Here, we mainly consider derivational morphology. Multiword expressions \eg, compound words are included as well. Words with different inflectional morphology are regarded as the same lemma form.}\cite{lieber2004morphology}, sense granularity and semantic relations \cite{sternefeld2013introduction}, motivated by linguistic and cognitive studies. Regarding syntactic categories, we focus on four \ie, parts of speech (\textbf{POS}) for target content words. Morphology aims at the number of morphemes (\textbf{nMorph}).
A sense inventory refers to the sense items in a dictionary, whose granularity influences the candidate sense listing for the target word and its sense annotation \cite{kilgarriff1997don}. We consider two aspects:
\begin{itemize}
    \item number of annotated ground-truth senses (\textbf{nGT});
    \item number of candidate senses, \ie, polysemy degree (\textbf{nPD}); 
\end{itemize}
    
To consider semantic interactions with other words, we utilize WordNet \cite{WordNet}, a semantic network to extract lexical relations. Specifically, we concentrate on the hyponym and synonymy relations. A word (or sense) is a hyponym of another if the first is more specific, denoting a subclass of the other. For example, \textit{table} is a hyponym of \textit{furniture}. Each word as a node in WordNet lies in a hyponym tree, where the depth implies the degree of specification, denoted as \textbf{dHypo}. Meanwhile, we also explore the size of the synonymy set (\textbf{dSyno}) into which the ground-truth sense falls. 

We perform linear regression analysis and conclude that most effects are significant as coefficients to the UE score, except for dSyno and ADV of POS. This is consistent with our result in Subection~\ref{sec:sr}. The summary of the linear regression is shown in Appendix~\ref{app:lrs}. \textcolor{black}{Afterwards, we design a controlled procedure to analyze and balance different effects. First, samples are drawn from all the test instances depending on some conditions, including nGT and POS. Afterward, we aggregate test data in one of three manners: \textit{instance} (I), \textit{lemma} (L), and \textit{sense} (S) and average the UE values for the instances with the same manner. I represents each occurrence of the target word, L considers words with different inflections (e.g., \textit{works} and \textit{worked}), and S targets words with the same ground-truth sense. The sampled data is then grouped into $N$ levels in terms of the values for the different effects in question. Finally, we calculate the mean UE score for each group and their corresponding T-test and p values. We heuristically set different choices of $N$ for different effects, considering the trade-off of level granularity and sample sparsity.} The p-value is expected to be lower than 5\%. The overall comparison is summarized in Table~\ref{tab:effects} with the number and value range of different levels in Table~\ref{tab:effect_level}.


\begin{table*}[t]
    \begin{center}
    \begin{tabular}{c|cc|ccc|ccc}
    \toprule
    \multirow{2}{*}{Effect} &
    \multirow{2}{*}{Condition} &
    \multirow{2}{*}{Agg.} &
    \multicolumn{3}{c|}{Uncertainty Estimation} & \multicolumn{3}{c}{Difference Significance}  \\
     & & & L1 & L2 & L3 & L1 $\leftrightarrow$ \text{L2} & L1 $\leftrightarrow$ \text{L3} & L2 $\leftrightarrow$ \text{L3} \\ 
    
    \midrule

\rowcolor[gray]{0.8}  & nGT=1, POS=NOUN & & 0.13 & 0.11 & 0.07 & \textbf{1.44e-2} & \textbf{1.35e-8} & \textbf{5e-4} \\ 
\rowcolor[gray]{0.8}     & nGT=1, POS=VERB & & 0.22 &0.19 & 0.13  & 7.61e-2 & \textbf{6.04e-4} & 6.6e-2 \\ 
      
\rowcolor[gray]{0.8}      & nGT=1, POS=ADJ & & 0.11 & 0.08 & 0.10 & \textbf{3.6e-2} & 4.21e-1 & 4.40e-1 \\ 
      
\rowcolor[gray]{0.8}  \multirow{-4}{*}{nMorph}    & nGT=1, POS=ADV & \multirow{-4}{*}{L}  & 0.11 & 0.06 & 0.02 & 7.6e-2 & \textbf{6.04e-4} & 6.60e-2 \\ 
      \midrule
nGT & - & I & 0.12 & 0.22 & - & \textbf{1.61e-22} & - & - \\ 
     
\rowcolor[gray]{0.8}     nPD & nGT=1 & L &0.04 & 0.16 & 0.22 & \textbf{6.22e-96} & \textbf{3.42e-135} & \textbf{5.01e-10} \\
     
     \midrule
      
       dHypo & nGT=1, POS=NOUN & L &  0.14 & 0.12 & 0.09 & \textbf{1.43e-2} & \textbf{1.91e-6} & \textbf{6e-3} \\
\rowcolor[gray]{0.8}       dSyno & nGT=1 & S & 0.14 & 0.14 & 0.14 & 5.55 & 5.38 & 5.67\\
      
        \bottomrule
          
    \end{tabular}
\end{center}
    \caption{Different uncertainty estimations (SMP) for different levels and corresponding difference significance (p values) of various effects involving morphology, inventory organization and semantic relations. Agg. means aggregation manners of the lemma (L), instance (I), and sense (S).}
    \label{tab:effects}
\end{table*}

\begin{table}[h!]
\small
    \centering
    \begin{tabular}{c|c|ccc}
        \toprule
        \multicolumn{2}{c|}{Effect} & L1  & L2  & L3  \\
        \midrule

 \rowcolor[gray]{0.8}        & number &   514 & 603 & 397 \\
   \rowcolor[gray]{0.8}  \multirow{-2}{*}{nMorph (N)}    & range & (0,1.67] & (1.67,2]  & (2,9] \\
        
         & number &   200 & 313 & 132 \\
     \multirow{-2}{*}{nMorph (V)}   & range & (0,2) & [2,2] & (2,6]  \\
        
   \rowcolor[gray]{0.8}        & number &   136 & 201 & 69 \\
  \rowcolor[gray]{0.8}     \multirow{-2}{*}{nMorph (A)}   & range & (0,1.30] & (1.30,2]  & (2,6] \\
        
         & number &   25 & 85 &36 \\
  \multirow{-2}{*}{nMorph (D)}      & range & (0,2] & [2,2] & (2,6] \\
        \midrule
    \rowcolor[gray]{0.8}       & number &   6913 & 340 & - \\
  \rowcolor[gray]{0.8}   \multirow{-2}{*}{nGT}     & range & 1 & >1  & - \\
        
         & number &   1145 & 963 & 463 \\
   \multirow{-2}{*}{nPD}     & range & (0,2] & (2,6]  & (6,50] \\
     
        \midrule
    \rowcolor[gray]{0.8}       & number &   729 & 666 & 340 \\
   \rowcolor[gray]{0.8}   \multirow{-2}{*}{dHypo}    & range & (1,6] & (6,9]  & (9,43] \\
        
        
         & number &  1109 & 1407 & 763 \\
     \multirow{-2}{*}{dSyno}   & range & (0,1] & (1,3] & (3,28]\\

        \bottomrule
    \end{tabular}
    \caption{The number and range of effects quantified into different levels for various effects.}
    \label{tab:effect_level}
\end{table}

\subsubsection{Syntactic Category and Morphology}

\begin{figure}[h!]
    \centering
    \includegraphics[width=1.0\linewidth]{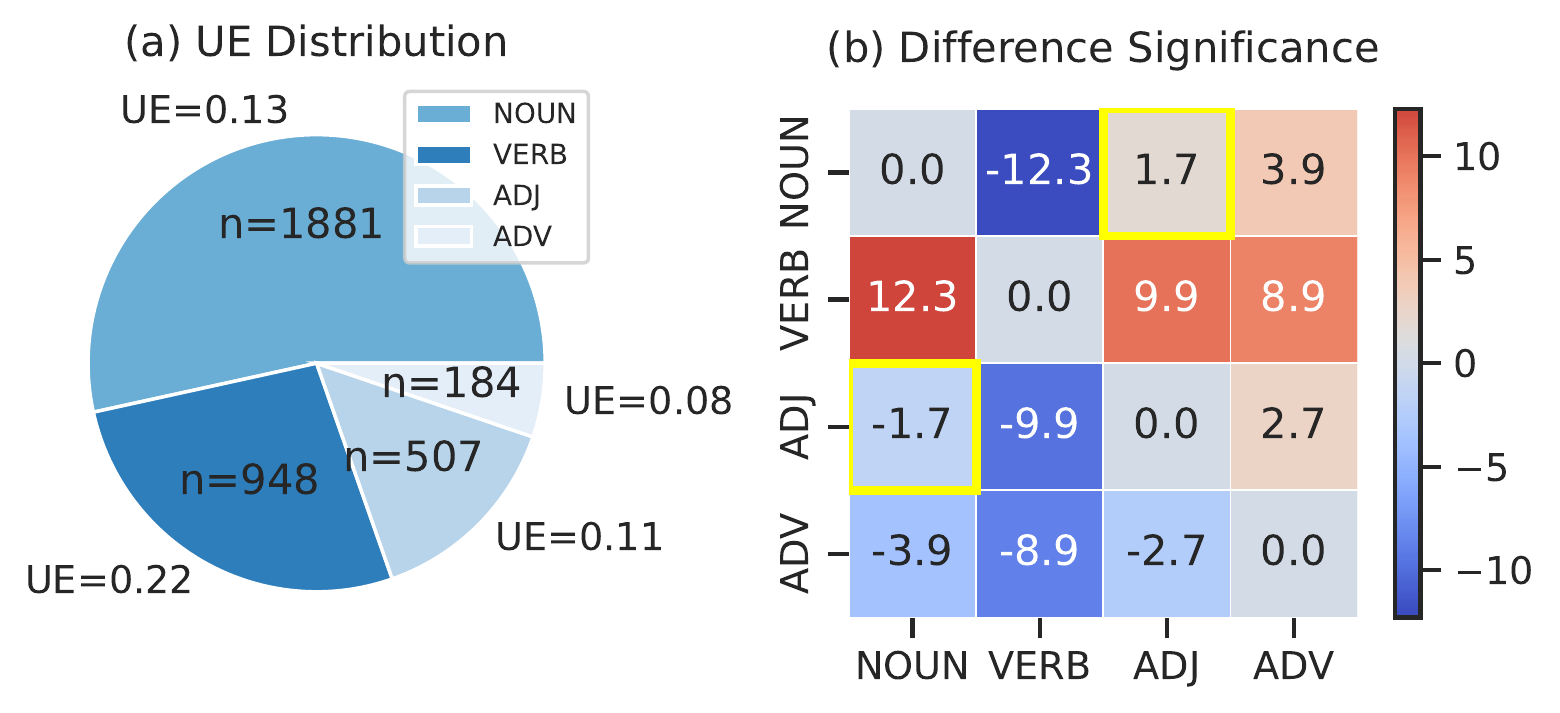}
    \caption{Averaged UE scores and numbers for instances aggregated by \textit{sense}, with different parts of speech (a) and the corresponding difference significance for each pair (b). The heatmap (b) shows the T-test values where a higher absolute value (grids with a deeper color) indicates a more significant difference. We highlight the grid with a corresponding p value larger than 5\%, implying no significant difference.}
    \label{fig:effect_pos}
\end{figure}

We show the averaged UE scores for instances with different POS and their corresponding T-test value in Figure~\ref{fig:effect_pos}. Except for the NOUN-ADJ pair, verbal instances are more significantly uncertain than NOUN or ADJ, while ADV has the least uncertainty. The result implies the senses of verbs are generally harder to determine than other categories, consistent with previous work \cite{ConSeC, DIBIMT}. This is reflected in Table~\ref{tab:all} and Figure~\ref{fig:effect_lemma}.

We further explore the effects of morphology in Table~\ref{tab:effects}. After extracting morphemes for each word using an off-line tool \footnote{\url{https://polyglot.readthedocs.io/en/latest}}, we count the number of morphemes (denoted as nMorph). Since words with different parts of speech may have distinct mechanisms of word formation rules, we split data according to POS before averaging their UE scores and calculating corresponding difference significance. It shows that generally, the more morphemes a word consists of, the more uncertain its semantics would be. This is expected from the perspective of derivational morphology since adding prefixes, or suffixes could specify the stem words and have a relatively predictable meaning. For example, ``V-ation'' indicates the action or process of the stem verb, \eg, education, memorization. According to T-test in Table~\ref{tab:effects}, UE scores of different levels for nouns are significantly distinct, while the difference is not so significant for other categories. \textcolor{black}{It is because the derivational nouns including compound words are more representative and productive than other categories. This can be demonstrated by the fact that nouns contain the highest number of morphemes as shown in Table~\ref{tab:effect_level}.}  

\subsubsection{Sense Granularity}
We first consider the number of ground-truth senses, \ie, nGT. During the annotation process, a not insignificant 5\% of the target words is labeled multiple senses \cite{ML-WSD}. This reflects the difficulty in choosing the most appropriate meaning, even for human annotators. Given their contexts, the semantics of these words are expected to be more uncertain, and our result is consistent with this fact. We control nGT to be 1 in the remaining evaluation to eliminate its influence.

Second, we study the effect of polysemy degree (the number of possible candidates), \ie, nPD. It shows that target words with a more significant polysemy degree tend to be more uncertain. It is intuitively understandable because words with more possible meanings are always commonplace and easily prone to semantic change, \eg, \textit{go}, \textit{play}. Furthermore, their sense descriptions in WordNet are more fine-grained, indistinguishable in some cases even for humans. However, words with less polysemy degrees, such as compound words, are more certain in various contexts. 

\subsubsection{Semantic relation} \label{sec:sr}
We discuss the effects of semantic relations for the target word in terms of WordNet. We first consider the hyponym relations, \ie, the depth in which a word node lies in the hyponym relation tree, as denoted by dHypo. Since nouns have clearer instances of hyponymy relation, we only consider this category. The results displayed in Table~\ref{tab:effects} show that instances with a deeper hyponym tend to own a certain meaning and the difference between each pair of levels is significant. \textcolor{black}{That indicates that more specific concepts have a more determinate disambiguation, which is intuitive.}    

Another semantic relation is synonymy, as represented by dSyno. The measurement reveals that instances among different levels of the number of synonyms do not differ from each other significantly. This implies that whether the ground-truth meaning has more neighbors with similar semantics has less impact on the decision of uncertainty.

\section{Conclusion}
We explore the uncertainty estimation for WSD. First, we compare various uncertainty scores. Then we choose SMP as the uncertainty indicator and examine to what extent a SOTA model captures data uncertainty and model uncertainty. Experiments demonstrate that the model estimates data uncertainty adequately but underestimates model uncertainty. We further explore effects that influence uncertainty estimation in the perspectives of morphology, inventory organization and semantic relations. We will integrate WSD with uncertainty estimation into downstream applications in the future. 

\section{Limitations}
Despite being easily adapted to current deep learning architectures, one concern about multiple-forward sampling methods is efficiency, since it has to repeat $T$ processes to evaluate uncertainty in the stage of inference. We leave efficient variants of sampling methods for future work.

Another glaring issue is the focus on only English. Different languages may have different effects on uncertainty estimation due to \eg, distinct forms of morphology. Thus, some conclusions may vary according to the language in question. We hope that follow-up works will refine and complement our insights on a more representative sample of natural languages.

\section{Ethics Statement}
We do not foresee any immediate negative ethical consequences of our research.

\section{Broader Impact Statement}
Knowing what we do not know, i.e., a well-calibrated uncertainty estimation, is fundamental for an AI-assisted application in the real world. In the area of word sense disambiguation, the ambiguity and vagueness inherent in lexical semantics require a model to represent and measure uncertainty effectively. Our work explores the combination of these two areas and hopes that it will provide an approach to understanding the characteristics of languages.

\section{Acknowledgements}
The authors thank the anonymous reviewers for their valuable comments and constructive feedback on the manuscript. We also thank Rui Fang for his discussions on the linear regression analysis. This work is supported by the 2018 National Major Program of Philosophy and Social Science Fund “Analyses and Researches of Classic Texts of Classical Literature Based on Big Data Technology” (18ZDA238) and Tsinghua University Initiative Scientific Research Program (2019THZWJC38). 

\bibliography{custom}

\begin{thebibliography}{50}
\expandafter\ifx\csname natexlab\endcsname\relax\def\natexlab#1{#1}\fi

\bibitem[{Agirre et~al.(2014)Agirre, L{\'o}pez~de Lacalle, and Soroa}]{UKB}
Eneko Agirre, Oier L{\'o}pez~de Lacalle, and Aitor Soroa. 2014.
\newblock Random walks for knowledge-based word sense disambiguation.
\newblock \emph{Computational Linguistics}, 40(1):57--84.

\bibitem[{Barba et~al.(2021)Barba, Procopio, and Navigli}]{ConSeC}
Edoardo Barba, Luigi Procopio, and Roberto Navigli. 2021.
\newblock Consec: Word sense disambiguation as continuous sense comprehension.
\newblock In \emph{Proceedings of the 2021 Conference on Empirical Methods in
  Natural Language Processing}, pages 1492--1503.

\bibitem[{Bevilacqua and Navigli(2019)}]{bevilacqua2019quasi}
Michele Bevilacqua and Roberto Navigli. 2019.
\newblock Quasi bidirectional encoder representations from transformers for
  word sense disambiguation.
\newblock In \emph{Proceedings of the International Conference on Recent
  Advances in Natural Language Processing (RANLP 2019)}, pages 122--131.

\bibitem[{Bevilacqua and Navigli(2020)}]{EWISER}
Michele Bevilacqua and Roberto Navigli. 2020.
\newblock Breaking through the 80\% glass ceiling: Raising the state of the art
  in word sense disambiguation by incorporating knowledge graph information.
\newblock In \emph{Proceedings of the 58th Annual Meeting of the Association
  for Computational Linguistics}, pages 2854--2864.

\bibitem[{Blevins and Zettlemoyer(2020)}]{BEM}
Terra Blevins and Luke Zettlemoyer. 2020.
\newblock Moving down the long tail of word sense disambiguation with gloss
  informed bi-encoders.
\newblock In \emph{Proceedings of the 58th Annual Meeting of the Association
  for Computational Linguistics}, pages 1006--1017.

\bibitem[{Calabrese et~al.(2021)Calabrese, Bevilacqua, and Navigli}]{EViLBERT}
Agostina Calabrese, Michele Bevilacqua, and Roberto Navigli. 2021.
\newblock Evilbert: Learning task-agnostic multimodal sense embeddings.
\newblock In \emph{Proceedings of the Twenty-Ninth International Conference on
  International Joint Conferences on Artificial Intelligence}, pages 481--487.

\bibitem[{Campolungo et~al.(2022)Campolungo, Martelli, Saina, and
  Navigli}]{DIBIMT}
Niccol{\`o} Campolungo, Federico Martelli, Francesco Saina, and Roberto
  Navigli. 2022.
\newblock Dibimt: A novel benchmark for measuring word sense disambiguation
  biases in machine translation.
\newblock In \emph{Proceedings of the 60th Annual Meeting of the Association
  for Computational Linguistics (Volume 1: Long Papers)}, pages 4331--4352.

\bibitem[{Conia and Navigli(2021)}]{ML-WSD}
Simone Conia and Roberto Navigli. 2021.
\newblock Framing word sense disambiguation as a multi-label problem for
  model-agnostic knowledge integration.
\newblock In \emph{Proceedings of the 16th Conference of the European Chapter
  of the Association for Computational Linguistics: Main Volume}, pages
  3269--3275.

\bibitem[{El-Yaniv et~al.(2010)}]{RCC}
Ran El-Yaniv et~al. 2010.
\newblock On the foundations of noise-free selective classification.
\newblock \emph{Journal of Machine Learning Research}, 11(5).

\bibitem[{Ferr{\'a}ndez et~al.(2006)Ferr{\'a}ndez, Roger, Ferr{\'a}ndez,
  Aguilar, and L{\'o}pez-Moreno}]{QA}
S~Ferr{\'a}ndez, Sandra Roger, Antonio Ferr{\'a}ndez, Antonia Aguilar, and
  Pilar L{\'o}pez-Moreno. 2006.
\newblock A new proposal of word sense disambiguation for nouns on a question
  answering system.
\newblock \emph{Advances in Natural Language Processing. Research in Computing
  Science}, 18:83--92.

\bibitem[{Folk and Morris(2003)}]{folk2003effects}
Jocelyn~R Folk and Robin~K Morris. 2003.
\newblock Effects of syntactic category assignment on lexical ambiguity
  resolution in reading: An eye movement analysis.
\newblock \emph{Memory \& Cognition}, 31:87--99.

\bibitem[{Gal and Ghahramani(2016)}]{gal2016dropout}
Yarin Gal and Zoubin Ghahramani. 2016.
\newblock Dropout as a bayesian approximation: Representing model uncertainty
  in deep learning.
\newblock In \emph{international conference on machine learning}, pages
  1050--1059. PMLR.

\bibitem[{Gal et~al.(2017)Gal, Islam, and Ghahramani}]{gal2017deep}
Yarin Gal, Riashat Islam, and Zoubin Ghahramani. 2017.
\newblock Deep bayesian active learning with image data.
\newblock In \emph{International Conference on Machine Learning}, pages
  1183--1192. PMLR.

\bibitem[{Geifman and El-Yaniv(2017)}]{geifman2017selective}
Yonatan Geifman and Ran El-Yaniv. 2017.
\newblock Selective classification for deep neural networks.
\newblock \emph{Advances in neural information processing systems}, 30.

\bibitem[{Gidiotis and Tsoumakas(2021)}]{gidiotis2021uncertainty}
Alexios Gidiotis and Grigorios Tsoumakas. 2021.
\newblock Uncertainty-aware abstractive summarization.
\newblock \emph{arXiv preprint arXiv:2105.10155}.

\bibitem[{Glushkova et~al.(2021)Glushkova, Zerva, Rei, and
  Martins}]{glushkova2021uncertainty}
Taisiya Glushkova, Chrysoula Zerva, Ricardo Rei, and Andr{\'e}~FT Martins.
  2021.
\newblock Uncertainty-aware machine translation evaluation.
\newblock In \emph{Findings of the Association for Computational Linguistics:
  EMNLP 2021}, pages 3920--3938.

\bibitem[{Gonzales et~al.(2017)Gonzales, Mascarell, and Sennrich}]{MT}
Annette~Rios Gonzales, Laura Mascarell, and Rico Sennrich. 2017.
\newblock Improving word sense disambiguation in neural machine translation
  with sense embeddings.
\newblock In \emph{Proceedings of the Second Conference on Machine
  Translation}, pages 11--19.

\bibitem[{Guo et~al.(2017)Guo, Pleiss, Sun, and
  Weinberger}]{guo2017calibration}
Chuan Guo, Geoff Pleiss, Yu~Sun, and Kilian~Q Weinberger. 2017.
\newblock On calibration of modern neural networks.
\newblock In \emph{International conference on machine learning}, pages
  1321--1330. PMLR.

\bibitem[{Hendrycks and Gimpel()}]{MP}
Dan Hendrycks and Kevin Gimpel.
\newblock A baseline for detecting misclassified and out-of-distribution
  examples in neural networks.
\newblock In \emph{International Conference on Learning Representations}.

\bibitem[{Heringer et~al.(1980)Heringer, Strecker, and
  Wimmer}]{heringer1980syntax}
Hans~J{\"u}rgen Heringer, Bruno Strecker, and Rainer Wimmer. 1980.
\newblock \emph{Syntax: Fragen, L{\"o}sungen, Alternativen}.
\newblock Fink.

\bibitem[{Houlsby et~al.(2011)Houlsby, Husz{\'a}r, Ghahramani, and
  Lengyel}]{houlsby2011bayesian}
Neil Houlsby, Ferenc Husz{\'a}r, Zoubin Ghahramani, and M{\'a}t{\'e} Lengyel.
  2011.
\newblock Bayesian active learning for classification and preference learning.
\newblock \emph{stat}, 1050:24.

\bibitem[{Hu and Liu(2011)}]{CR}
Shangfeng Hu and Chengfei Liu. 2011.
\newblock Incorporating coreference resolution into word sense disambiguation.
\newblock In \emph{International Conference on Intelligent Text Processing and
  Computational Linguistics}, pages 265--276. Springer.

\bibitem[{Huang et~al.(2019)Huang, Sun, Qiu, and Huang}]{GBERT}
Luyao Huang, Chi Sun, Xipeng Qiu, and Xuan-Jing Huang. 2019.
\newblock Glossbert: Bert for word sense disambiguation with gloss knowledge.
\newblock In \emph{Proceedings of the 2019 Conference on Empirical Methods in
  Natural Language Processing and the 9th International Joint Conference on
  Natural Language Processing (EMNLP-IJCNLP)}, pages 3509--3514.

\bibitem[{Kenton and Toutanova(2019)}]{bert}
Jacob Devlin Ming-Wei~Chang Kenton and Lee~Kristina Toutanova. 2019.
\newblock Bert: Pre-training of deep bidirectional transformers for language
  understanding.
\newblock In \emph{Proceedings of NAACL-HLT}, pages 4171--4186.

\bibitem[{Kilgarriff(1997)}]{kilgarriff1997don}
Adam Kilgarriff. 1997.
\newblock I don’t believe in word senses.
\newblock \emph{Computers and the Humanities}, 31(2):91--113.

\bibitem[{Kochkina and Liakata(2020)}]{kochkina2020estimating}
Elena Kochkina and Maria Liakata. 2020.
\newblock Estimating predictive uncertainty for rumour verification models.
\newblock In \emph{Proceedings of the 58th Annual Meeting of the Association
  for Computational Linguistics}, pages 6964--6981.

\bibitem[{Lieber(2004)}]{lieber2004morphology}
Rochelle Lieber. 2004.
\newblock \emph{Morphology and lexical semantics}, volume 104.
\newblock Cambridge University Press.

\bibitem[{Loureiro and Jorge(2019)}]{loureiro2019language}
Daniel Loureiro and Alipio Jorge. 2019.
\newblock Language modelling makes sense: Propagating representations through
  wordnet for full-coverage word sense disambiguation.
\newblock In \emph{Proceedings of the 57th Annual Meeting of the Association
  for Computational Linguistics}, pages 5682--5691.

\bibitem[{Maru et~al.(2022)Maru, Conia, Bevilacqua, and Navigli}]{42D}
Marco Maru, Simone Conia, Michele Bevilacqua, and Roberto Navigli. 2022.
\newblock Nibbling at the hard core of word sense disambiguation.
\newblock In \emph{Proceedings of the 60th Annual Meeting of the Association
  for Computational Linguistics (Volume 1: Long Papers)}, pages 4724--4737.

\bibitem[{Miller et~al.(1990)Miller, Beckwith, Fellbaum, Gross, and
  Miller}]{WordNet}
George~A Miller, Richard Beckwith, Christiane Fellbaum, Derek Gross, and
  Katherine~J Miller. 1990.
\newblock Introduction to wordnet: An on-line lexical database.
\newblock \emph{International journal of lexicography}, 3(4):235--244.

\bibitem[{Miller et~al.(1994)Miller, Chodorow, Landes, Leacock, and
  Thomas}]{semcor}
George~A Miller, Martin Chodorow, Shari Landes, Claudia Leacock, and Robert~G
  Thomas. 1994.
\newblock Using a semantic concordance for sense identification.
\newblock In \emph{Human Language Technology: Proceedings of a Workshop held at
  Plainsboro, New Jersey, March 8-11, 1994}.

\bibitem[{Moro et~al.(2014)Moro, Raganato, and Navigli}]{Babelfy}
Andrea Moro, Alessandro Raganato, and Roberto Navigli. 2014.
\newblock Entity linking meets word sense disambiguation: a unified approach.
\newblock \emph{Transactions of the Association for Computational Linguistics},
  2:231--244.

\bibitem[{Navigli(2009)}]{2009survey}
Roberto Navigli. 2009.
\newblock Word sense disambiguation: A survey.
\newblock \emph{ACM computing surveys (CSUR)}, 41(2):1--69.

\bibitem[{Navigli and Ponzetto(2012)}]{BabelNet}
Roberto Navigli and Simone~Paolo Ponzetto. 2012.
\newblock Babelnet: The automatic construction, evaluation and application of a
  wide-coverage multilingual semantic network.
\newblock \emph{Artificial intelligence}, 193:217--250.

\bibitem[{Neal(2012)}]{neal2012bayesian}
Radford~M Neal. 2012.
\newblock \emph{Bayesian learning for neural networks}, volume 118.
\newblock Springer Science \& Business Media.

\bibitem[{Penha and Hauff(2021)}]{penha2021calibration}
Gustavo Penha and Claudia Hauff. 2021.
\newblock On the calibration and uncertainty of neural learning to rank models
  for conversational search.
\newblock In \emph{Proceedings of the 16th Conference of the European Chapter
  of the Association for Computational Linguistics: Main Volume}, pages
  160--170.

\bibitem[{Qi et~al.(2020)Qi, Zhang, Zhang, Bolton, and Manning}]{stanza}
Peng Qi, Yuhao Zhang, Yuhui Zhang, Jason Bolton, and Christopher~D Manning.
  2020.
\newblock Stanza: A python natural language processing toolkit for many human
  languages.
\newblock In \emph{Proceedings of the 58th Annual Meeting of the Association
  for Computational Linguistics: System Demonstrations}, pages 101--108.

\bibitem[{Raganato et~al.(2017)Raganato, Camacho-Collados, and
  Navigli}]{evaluation}
Alessandro Raganato, Jose Camacho-Collados, and Roberto Navigli. 2017.
\newblock Word sense disambiguation: A unified evaluation framework and
  empirical comparison.
\newblock In \emph{Proceedings of the 15th Conference of the European Chapter
  of the Association for Computational Linguistics: Volume 1, Long Papers},
  pages 99--110.

\bibitem[{Sternefeld and Zimmermann(2013)}]{sternefeld2013introduction}
Wolfgang Sternefeld and Thomas~Ede Zimmermann. 2013.
\newblock \emph{Introduction to Semantics: An Essential Guide to the
  Composition of Meaning (Mouton Textbook)}.
\newblock De Gruyter Mouton.

\bibitem[{Stutz(2022)}]{stutz2022understanding}
David Stutz. 2022.
\newblock Understanding and improving robustness and uncertainty estimation in
  deep learning.
\newblock \emph{Saarl{\"a}ndische Universit{\"a}ts-und Landesbibliothek}.

\bibitem[{Szegedy et~al.(2016)Szegedy, Vanhoucke, Ioffe, Shlens, and
  Wojna}]{LabelSmooth}
Christian Szegedy, Vincent Vanhoucke, Sergey Ioffe, Jon Shlens, and Zbigniew
  Wojna. 2016.
\newblock Rethinking the inception architecture for computer vision.
\newblock In \emph{Proceedings of the IEEE conference on computer vision and
  pattern recognition}, pages 2818--2826.

\bibitem[{Tripodi and Navigli(2019)}]{GT}
Rocco Tripodi and Roberto Navigli. 2019.
\newblock Game theory meets embeddings: a unified framework for word sense
  disambiguation.
\newblock In \emph{Proceedings of the 2019 Conference on Empirical Methods in
  Natural Language Processing and the 9th International Joint Conference on
  Natural Language Processing (EMNLP-IJCNLP)}, pages 88--99.

\bibitem[{Varshney and Baral(2023)}]{varshney2023post}
Neeraj Varshney and Chitta Baral. 2023.
\newblock Post-abstention: Towards reliably re-attempting the abstained
  instances in qa.
\newblock \emph{arXiv preprint arXiv:2305.01812}.

\bibitem[{Varshney et~al.(2022)Varshney, Mishra, and
  Baral}]{varshney2022towards}
Neeraj Varshney, Swaroop Mishra, and Chitta Baral. 2022.
\newblock Towards improving selective prediction ability of nlp systems.
\newblock In \emph{Proceedings of the 7th Workshop on Representation Learning
  for NLP}, pages 221--226.

\bibitem[{Vazhentsev et~al.(2022)Vazhentsev, Kuzmin, Shelmanov, Tsvigun,
  Tsymbalov, Fedyanin, Panov, Panchenko, Gusev, Burtsev
  et~al.}]{vazhentsev2022uncertainty}
Artem Vazhentsev, Gleb Kuzmin, Artem Shelmanov, Akim Tsvigun, Evgenii
  Tsymbalov, Kirill Fedyanin, Maxim Panov, Alexander Panchenko, Gleb Gusev,
  Mikhail Burtsev, et~al. 2022.
\newblock Uncertainty estimation of transformer predictions for
  misclassification detection.
\newblock In \emph{Proceedings of the 60th Annual Meeting of the Association
  for Computational Linguistics (Volume 1: Long Papers)}, pages 8237--8252.

\bibitem[{Weaver(1952)}]{weaver1952translation}
Warren Weaver. 1952.
\newblock Translation.
\newblock In \emph{Proceedings of the Conference on Mechanical Translation}.

\bibitem[{Xin et~al.(2021)Xin, Tang, Yu, and Lin}]{RPP}
Ji~Xin, Raphael Tang, Yaoliang Yu, and Jimmy Lin. 2021.
\newblock The art of abstention: Selective prediction and error regularization
  for natural language processing.
\newblock In \emph{Proceedings of the 59th Annual Meeting of the Association
  for Computational Linguistics and the 11th International Joint Conference on
  Natural Language Processing (Volume 1: Long Papers)}, pages 1040--1051.

\bibitem[{Yarin(2016)}]{galthesis}
Gal Yarin. 2016.
\newblock Uncertainty in deep learning.
\newblock \emph{University of Cambridge, Cambridge}.

\bibitem[{Zhu et~al.(2008)Zhu, Wang, Yao, and Tsou}]{zhu2008active}
Jingbo Zhu, Huizhen Wang, Tianshun Yao, and Benjamin~K Tsou. 2008.
\newblock Active learning with sampling by uncertainty and density for word
  sense disambiguation and text classification.
\newblock In \emph{Proceedings of the 22nd International Conference on
  Computational Linguistics (Coling 2008)}, pages 1137--1144.

\bibitem[{Zwillinger and Kokoska(1999)}]{zwillinger1999crc}
Daniel Zwillinger and Stephen Kokoska. 1999.
\newblock \emph{CRC standard probability and statistics tables and formulae}.
\newblock Crc Press.

\end{thebibliography}
\bibliographystyle{acl_natbib}
\clearpage

\appendix

\section{Appendix}
\label{sec:appendix}

\subsection{Distribution of UE scores}
\label{app:distribution}
\begin{figure}[h!]
    \centering
    \includegraphics[width=1.0\linewidth]{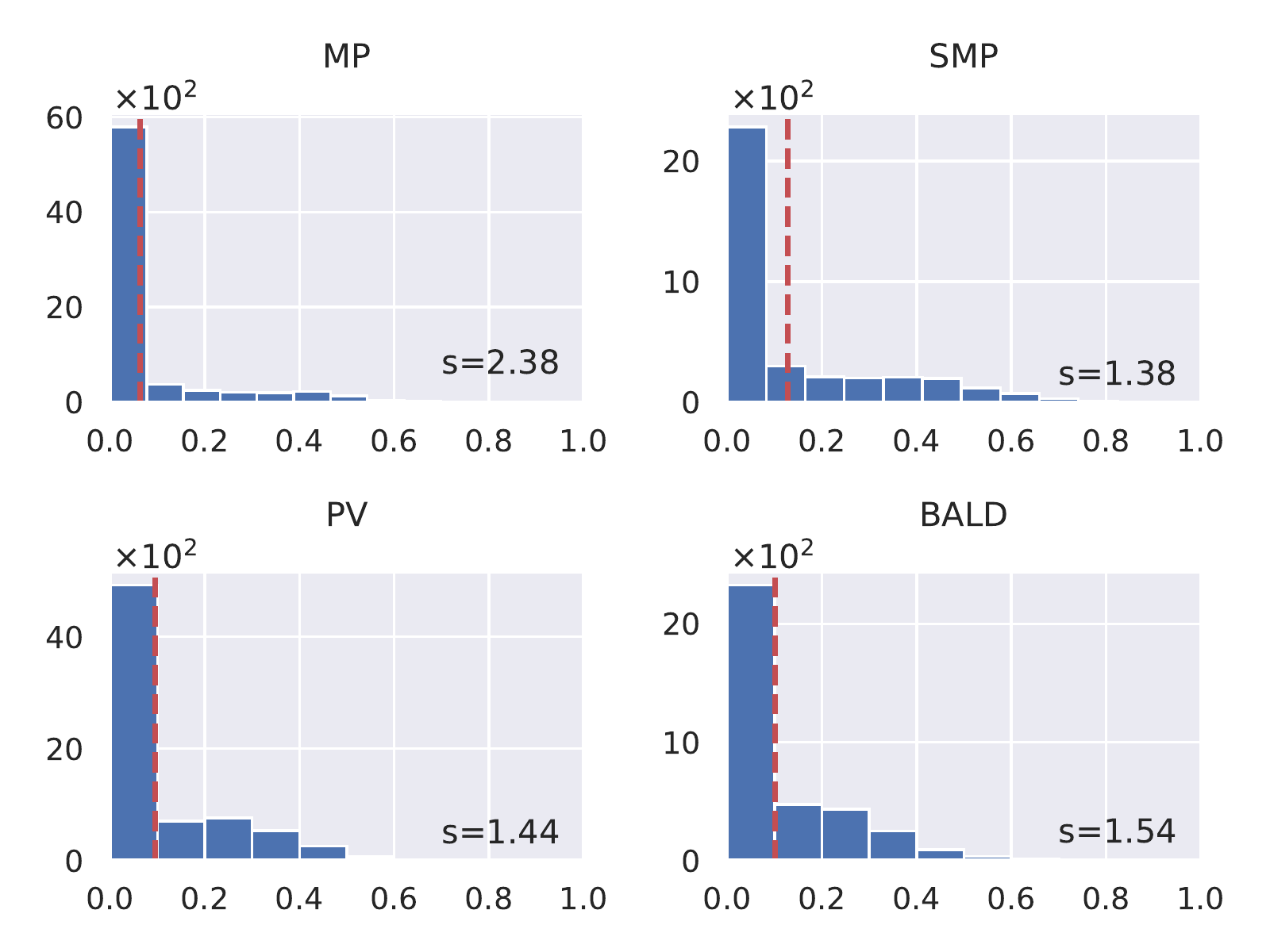}
    \caption{The distribution of four UE scores on all the test samples. The averaged value is indicated by a red dotted line. We calculate the sample skewness for each score as well.}
    \label{fig:com_dist_all}
\end{figure}

\begin{figure}[h!]
    \centering
    \includegraphics[width=1.0\linewidth]{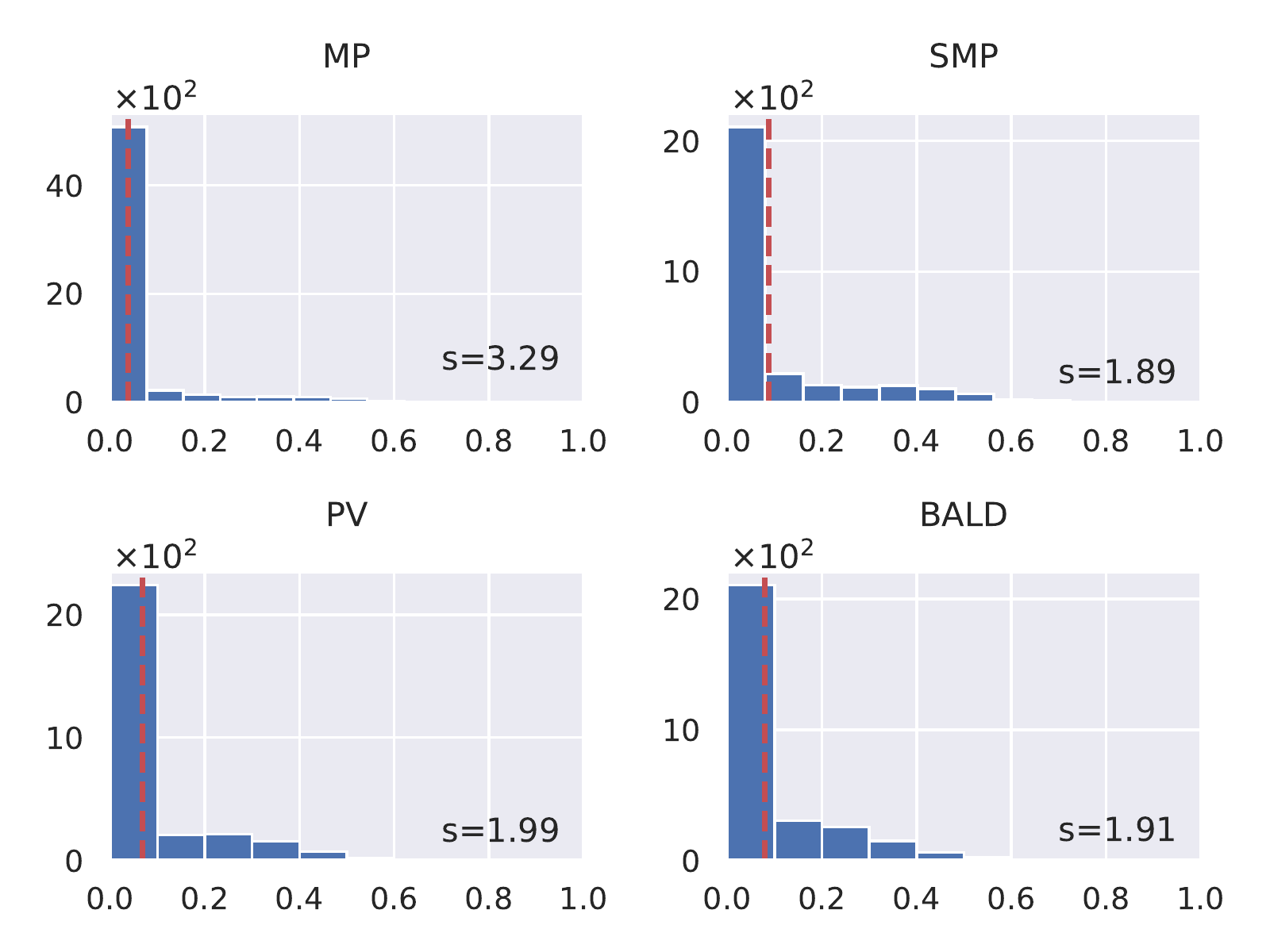}
    \caption{UE distribution on well-classified samples.}
    \label{fig:com_dist_right}
\end{figure}

We illustrate the distribution of UE scores, \ie, MP, SMP, PV and BALD for all the test samples in Figure~\ref{fig:com_dist_all} and samples that are correctly predicted in Figure~\ref{fig:com_dist_right}. We assume samples that the model could accurately predict are easy and thus have a more certain meaning. Although SMP is not so long-tailed as MP in the case of correctly predicted samples, we do not expect a metric ``overconfident'' in all the cases, especially in the misclassified instances.

\subsection{Other Scores for Data Uncertainty}
\label{app:du}
\begin{figure}[h!]
    \centering
    \includegraphics[width=1.0\linewidth]{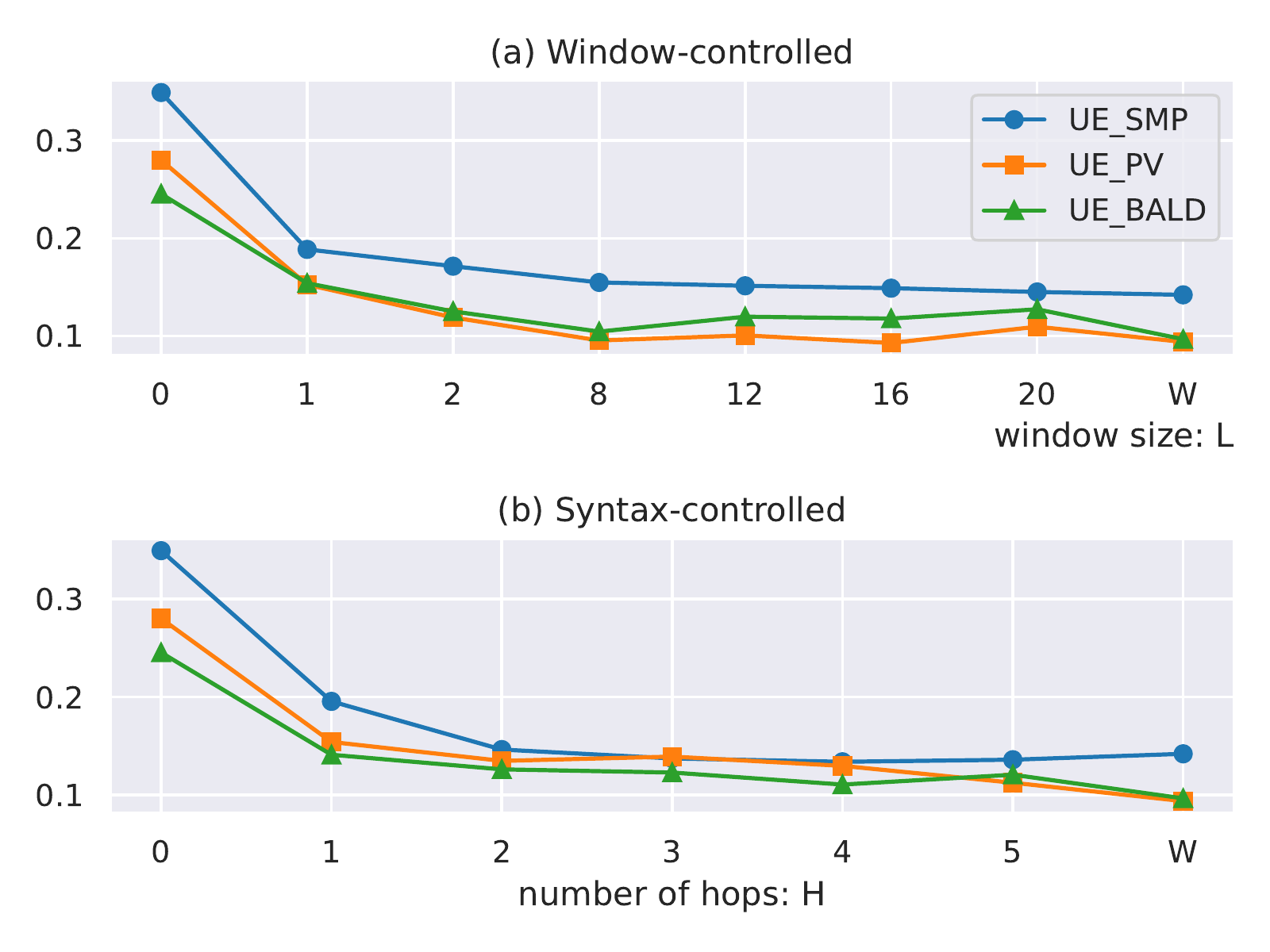}
    \caption{UE scores (SMP, PV, and BALD) vary depending on the range of context for (a) window-controlled setting and (b) syntax-controlled setting.}
    \label{fig:du_other}
\end{figure}

We display the other two sample-based scores PV and BALD, in comparison with SMP in two data uncertainty scenarios in Figure~\ref{fig:du_other}. SMP has a higher uncertain score than the other two, especially in the more sparse context (\eg, $L$ = 0), as we expected.

\subsection{Other Scores for Model Uncertainty}
\label{app:mu}
\begin{figure}[h!]
    \centering
    \includegraphics[width=1.0\linewidth]{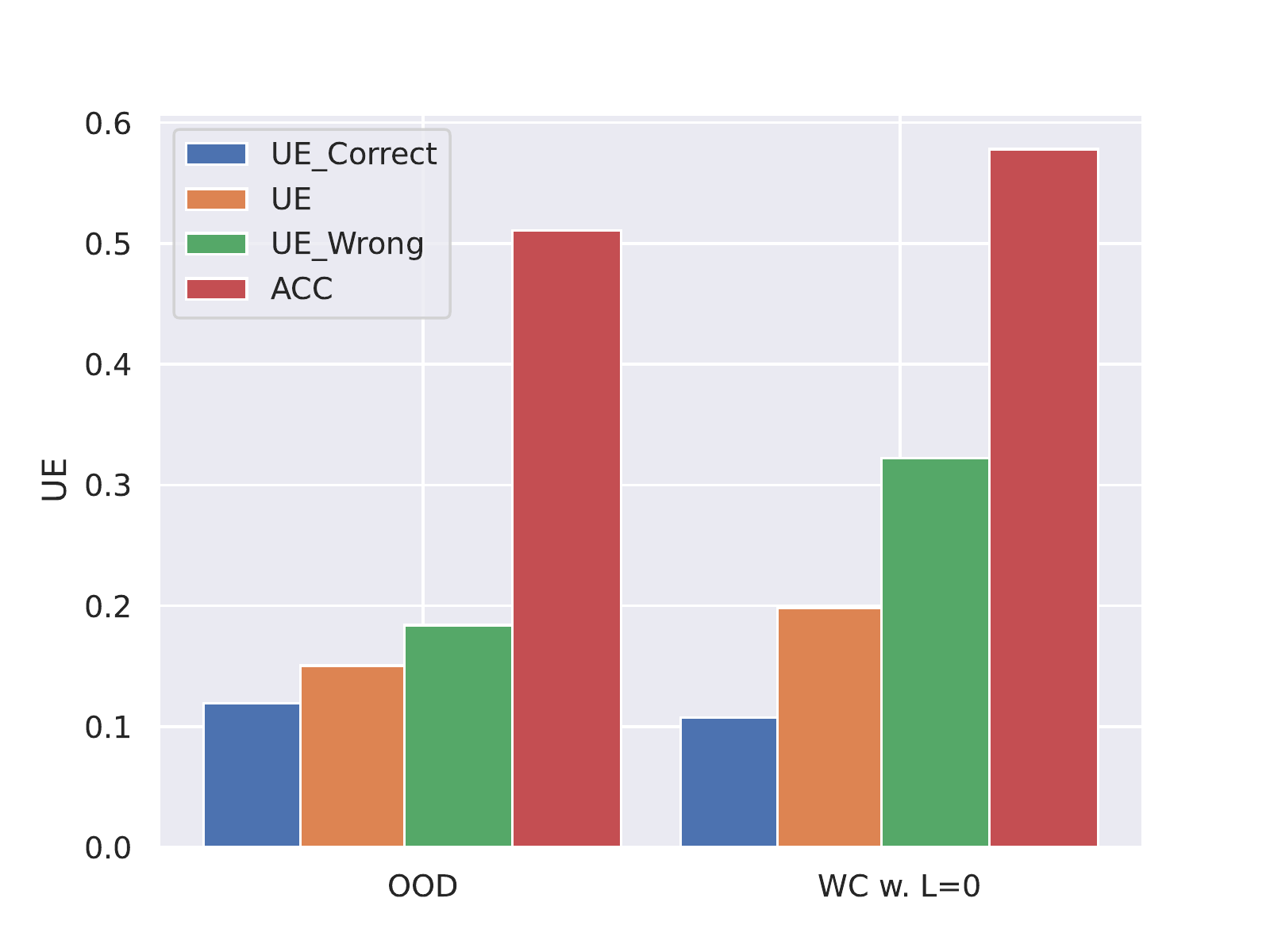}
    \caption{Uncertainty (MP) and accuracy scores for model uncertainty (OOD) and data uncertainty (controlled context) scenarios. We use window-controlled UE with \textit{L}=0 (WC w. \textit{L}=0). It is evaluated in all the data instances and wrongly (UE\_Wrong) or correctly (UE\_Correct) classified instances.}
    \label{fig:OOD_MP}
\end{figure}

\begin{figure}[h!]
    \centering
    \includegraphics[width=1.0\linewidth]{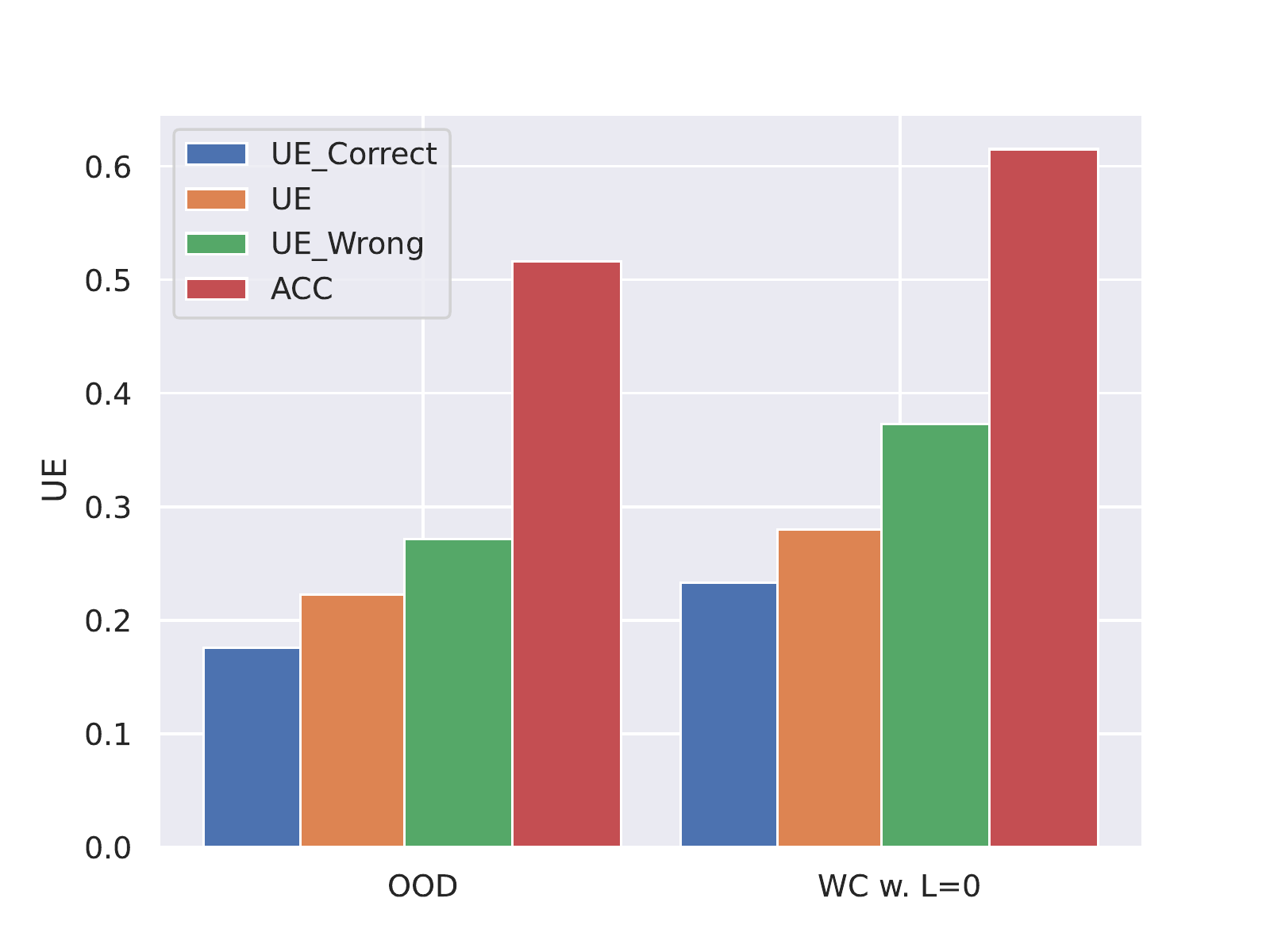}
    \caption{Uncertainty (PV) and accuracy scores for model uncertainty (OOD) and data uncertainty (controlled context) scenarios.}
    \label{fig:OOD_PV}
\end{figure}

\begin{figure}[h!]
    \centering
    \includegraphics[width=1.0\linewidth]{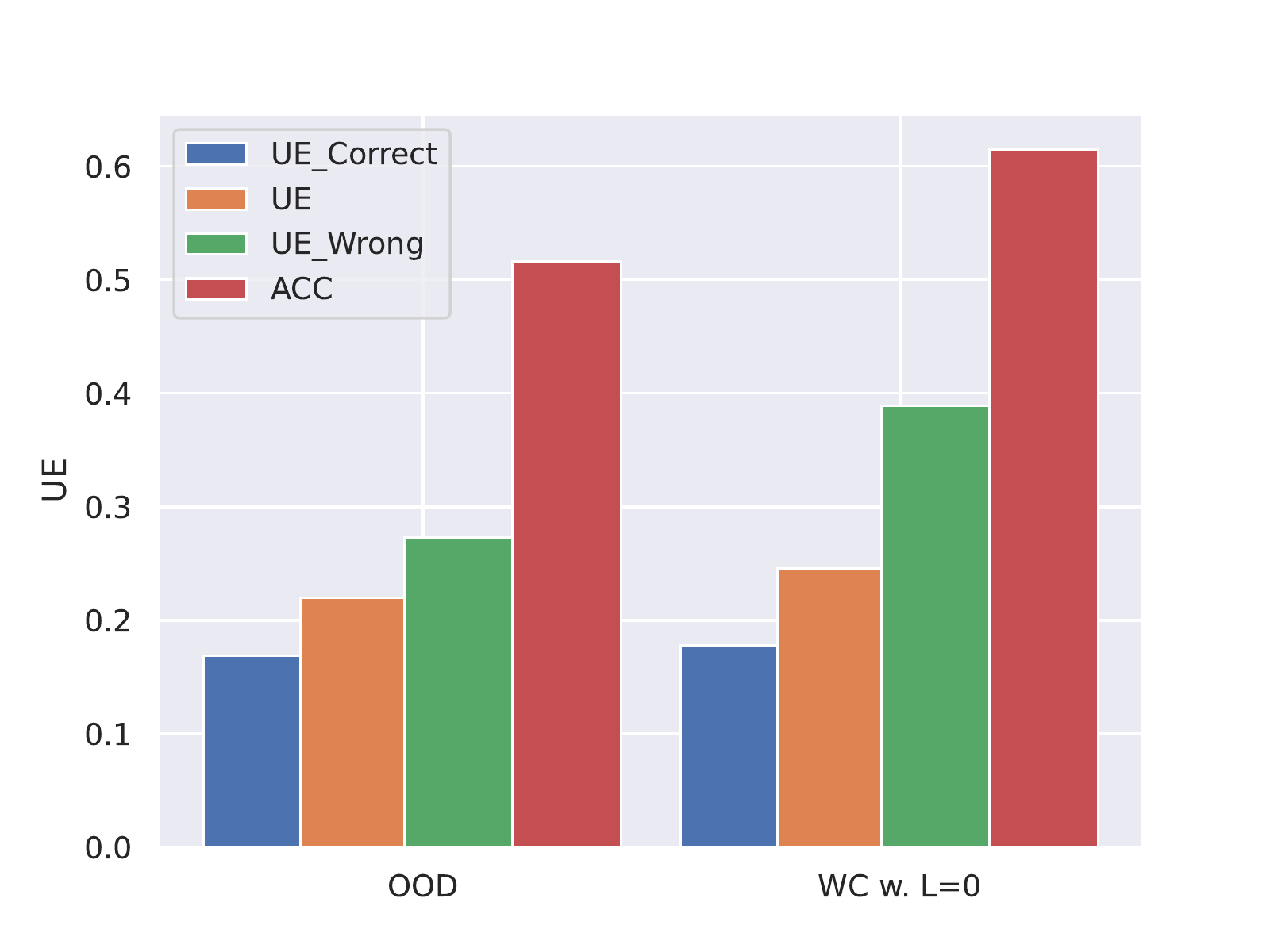}
    \caption{Uncertainty (BALD) and accuracy scores for model uncertainty (OOD) and data uncertainty (controlled context) scenarios.}
    \label{fig:OOD_BALD}
\end{figure}

We illustrate the other three UE scores (MP, PV and BALD) and accuracy for the scenario of model uncertainty compared with the least uncertain case for data uncertainty ($L$=0) in Figure~\ref{fig:OOD_MP}, Figure~\ref{fig:OOD_PV} and Figure~\ref{fig:OOD_BALD}, respectively. The conclusion that UE scores underestimate model uncertainty is similar to that of MP.

\subsection{Linear Regression Analysis}
\label{app:lrs}
\begin{figure}[h!]
    \centering
    \includegraphics[width=0.92\linewidth]{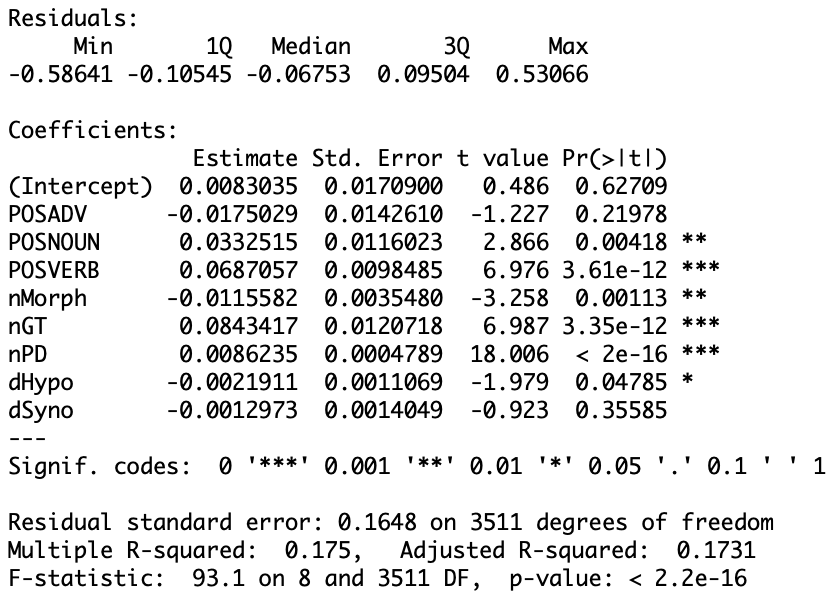}
    \caption{Linear regression model predicting the UE score (SMP) by various effects.}
    \label{fig:LR}
\end{figure}

Figure~\ref{fig:LR} reports all the effects and corresponding coefficients and p-values of the linear regression model described in Subsection~\ref{sec:effects}.

\end{document}